\documentclass[10pt,a4paper,logo]{paper}
\usepackage{fontawesome5}

\usepackage{kotex}
\usepackage{placeins}
\usepackage{afterpage}
\usepackage{float}

\usepackage[all]{hypcap}

\usepackage[authoryear, round]{natbib}

\usepackage{nicefrac}       %
\usepackage{multirow}
\usepackage{cleveref}
\usepackage{dsfont}
\usepackage{makecell}
\usepackage{subcaption}
\usepackage{bbm}
\usepackage{wrapfig}
\usepackage{needspace}
\usepackage{mathtools,xparse}
\usepackage{array}
\usepackage{colortbl}
\usepackage{arydshln}
\usepackage{scalerel}
\usepackage{tablefootnote}
\usepackage{tocloft}
\usepackage{adjustbox}
\usepackage{setspace}
\usepackage[most,skins,theorems]{tcolorbox}

\newtcolorbox{promptbox}{
  colback=gray!5,
  colframe=gray!70,
  boxrule=0.8pt,
  arc=2mm,
  left=10pt,
  right=10pt,
  top=10pt,
  bottom=10pt,
  width=\linewidth,
  fontupper=\ttfamily\footnotesize,
  enhanced,
}

\newboolean{showsection}
\setboolean{showsection}{true}
\makeatletter
\@namedef{ver@everyshi.sty}{}
\makeatother

\definecolor{custom_green}{rgb}{0.0, 0.5, 0.0}
\definecolor{custom_red}{rgb}{1.0, 0.01, 0.24}
\definecolor{custom_blue}{HTML}{C9DAF7}
\definecolor{custom_purple}{HTML}{D9D1E9}
\definecolor{title_blue}{HTML}{204899}
\definecolor{cite_blue}{HTML}{044dc1}
\definecolor{cite_purple}{HTML}{7406a7}

\hypersetup{
    colorlinks = true,
    citecolor = {cite_blue},
    linkcolor = {cite_purple},
    urlcolor = {cite_purple},
}

\def\Eqref#1{Equation~\ref{#1}}

\definecolor{blanchedalmond}{rgb}{1.0, 0.92, 0.8}
\definecolor{carmine}{rgb}{0.59, 0.0, 0.09}
\definecolor{lightblue}{rgb}{0.22,0.45,0.70}%

\renewcommand{\mathbf}{\boldsymbol}

\makeatletter
\def\Ddots{\mathinner{\mkern1mu\raise\p@
\vbox{\kern7\p@\hbox{.}}\mkern2mu
\raise4\p@\hbox{.}\mkern2mu\raise7\p@\hbox{.}\mkern1mu}}
\makeatother

\numberwithin{equation}{section}

\definecolor{amaranth}{rgb}{0.9, 0.17, 0.31}
\definecolor{antiquebrass}{rgb}{0.8, 0.58, 0.46}
\definecolor{antiquefuchsia}{rgb}{0.57, 0.36, 0.51}
\definecolor{chromeyellow}{rgb}{0.31, 0.47, 0.26}

\newcommand{\1}{\mathds 1}

\DeclareRobustCommand{\figstep}[1]{{\normalfont\textcircled{#1}}}
\newcommand{\steptransition}[2]{\mbox{\figstep{#1}\,$\rightarrow$\,\figstep{#2}}}

\usepackage{amsmath,amsfonts,bm}

\def\eqref#1{equation~\ref{#1}}
\def\Eqref#1{Equation~\ref{#1}}

\def\1{\bm{1}}

\DeclareMathAlphabet{\mathsfit}{\encodingdefault}{\sfdefault}{m}{sl}
\SetMathAlphabet{\mathsfit}{bold}{\encodingdefault}{\sfdefault}{bx}{n}

\makeatletter
\def\mathcolor#1#{\@mathcolor{#1}}
\def\@mathcolor#1#2#3{%
  \protect\leavevmode
  \begingroup
    \color#1{#2}#3%
  \endgroup
}
\makeatother

\Crefformat{equation}{#2Eq.\;(#1)#3}
\Crefformat{figure}{#2Figure #1#3}
\Crefformat{assumption}{#2Assumption #1#3}
\Crefname{assumption}{Assumption}{Assumptions}

\usepackage{crossreftools}
\makeatletter
\renewcommand\footnoterule{%
  \kern 15\p@
  \hrule \@width 2in \kern 2.6\p@
  \vspace{4pt}
}
\makeatother

\title{\textsc{ResidualQuant}: KV Cache Quantization for Looped Transformers with 2-Bit Residuals}

\reportnumber{}

\author[1]{Heejun Kim}
\author[2]{Junyoung Lee}
\author[3]{SangLyul Cho}
\author[1]{Dongsu Han}
\author[1]{Insu Han}
\author[1,$\dagger$]{Sehoon Kim}
\affil[1]{KAIST}
\affil[2]{Yonsei University}
\affil[3]{Seoul National University}
\correspondingauthor{$\dagger$ Corresponding author: Sehoon Kim.}

\begin{abstract}
\textbf{Abstract:}
Looped Transformers improve parameter efficiency by applying shared Transformer blocks over recurrent loops, increasing computational depth without increasing the parameter count.
However, KV cache memory still scales with loop count, becoming a key memory bottleneck that limits batch size and inference throughput.
KV cache quantization can alleviate this bottleneck, but existing methods often suffer substantial accuracy degradation at aggressive low-precision regimes.
We observe that looped Transformers offer a unique opportunity: KV states across loops are highly similar.
Based on this observation, we propose \textsc{ResidualQuant}, which uses the final-loop KV states as a reference and represents the remaining loops with low-precision residuals.
Our method further combines least-square scaling and rotation-based quantization, as well as loop-wise mixed precision, to enable accurate quantization down to INT2 while retaining efficient reconstruction.
Across multiple looped Transformer models and mathematical reasoning and code generation benchmarks, \textsc{ResidualQuant} consistently improves the accuracy--memory tradeoff over state-of-the-art rotation-based KV quantization.
In particular, our method retains accuracy close to BF16 under mixed-precision settings while reducing theoretical KV storage by 80.7\%, achieving up to 13.0\% higher accuracy than the rotation-based baseline at nearly the same memory budget.
On an RTX 5090, reduced KV memory traffic improves fixed-batch decode throughput by up to $3.66\times$, while the smaller memory footprint enables up to $8\times$ larger batches, improving peak throughput by up to $5.38\times$.

\par\smallskip
\noindent\faGlobe\enspace\textbf{Project page:} \url{https://seal-kaist.github.io/projects/ResidualQuant}
\end{abstract}

\begin{document}

\maketitle

\begin{figure}[H]
\centering
\includegraphics[width=\linewidth]{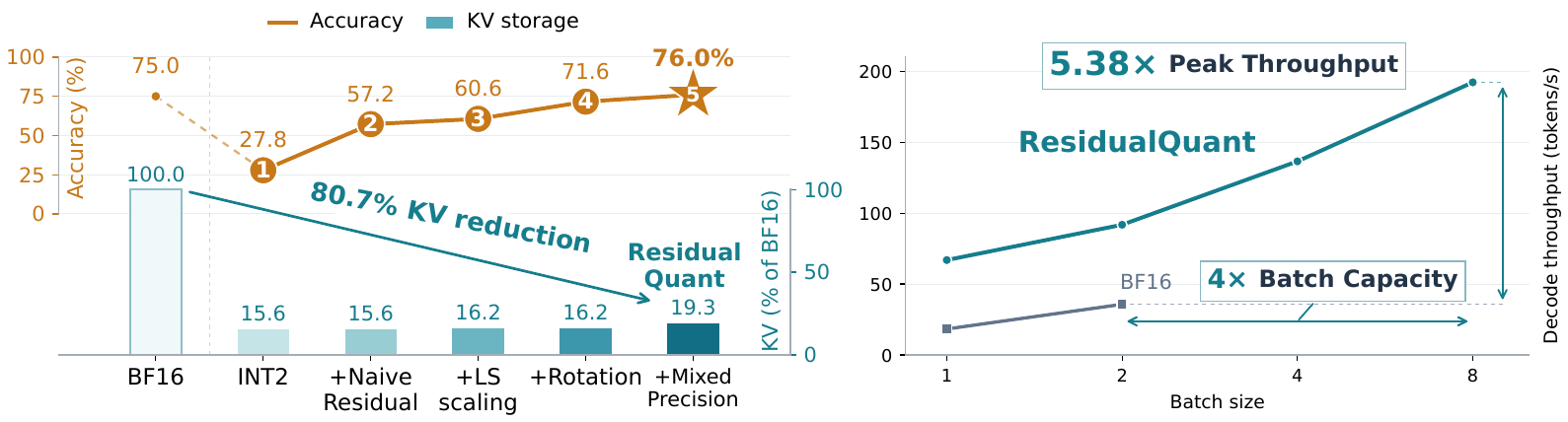}
\caption{\textbf{\textsc{ResidualQuant} progressively recovers BF16-level accuracy while reducing KV storage and improving decode throughput.} Results on Ouro-1.4B with group size $g=32$ in every quantized loop. Left: Starting from \figstep{1} direct INT2 quantization, we successively introduce \figstep{2} Last-loop residual quantization,  \figstep{3} least-square scaling,  \figstep{4} rotation, and \figstep{5} mixed precision to achieve \textsc{ResidualQuant}. Accuracy increases from 27.8\% to 76.0\%, matching the BF16 baseline of 75.0\%, while KV cache storage is reduced by 80.7\%. Right: At 16k context on an RTX 5090, \textsc{ResidualQuant} achieves $5.38\times$ the peak decode throughput and $4\times$ the largest feasible batch size compared to the BF16 counterpart.
}
\label{fig:acc-throughput}
\end{figure}

\section{Introduction}
Recursive or looped Transformers refine latent representations by repeatedly applying shared modules, increasing computational depth without increasing the parameter count \citep{dehghani2019universal,saunshi2025latent,suleymanzade2026loopedflows,wang2026recurtrace}.
Models such as Huginn \citep{geiping2025huginn} and Ouro \citep{zhu2025ouro} exploit this structure through adjustable recurrence or adaptive depth allocation, requiring less weight storage than models with distinct parameters at the same computational depth.
However, each loop produces distinct keys and values, making KV cache memory grow with recurrent depth, in addition to context length and batch size, despite weight sharing.
This limits the extent to which weight savings translate into larger batches and higher throughput, making KV cache memory reduction an important challenge for realizing the full efficiency benefits of looped Transformers.

To reduce storage overhead, prior work has explored sharing or reusing KV states across recurrences.
MoR \citep{bae2025mor} reuses KV states from the first recursion, PLT \citep{wu2025plt} combines first-loop global KV with local attention in subsequent loops, and MELT \citep{vendrell2026melt} updates shared KV states through learned gates.
While these approaches reduce storage, sharing KV states across loops can discard loop-specific information and degrade accuracy when representations differ across loops.

\begin{figure}[!t]
\centering
\includegraphics[width=\linewidth]{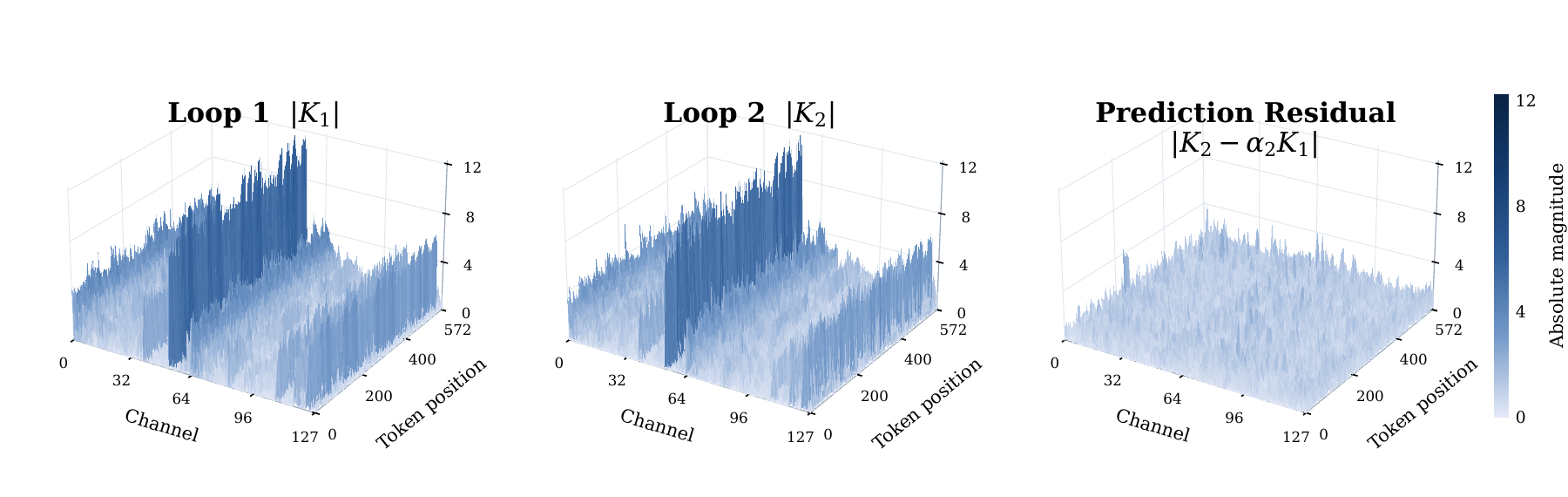}
\caption{\textbf{Post-RoPE key magnitudes in Ouro-1.4B.} Left: Loop 1, $|K_1|$. Middle: Loop 2, $|K_2|$. Right: the residual after Least Square scaling of Loop 1's keys, $|K_2-\alpha_2K_1|$. All panels use the same height and color scales. Visualization details and last-loop anchor comparisons are provided in Appendix~\ref{app:e1-transitions}.
}
\label{fig:e1_loop1_loop2}
\end{figure}

KV quantization offers an alternative approach that preserves distinct KV states for each loop while reducing their storage precision.
Because KV memory grows with recurrent depth, aggressive low-precision quantization is particularly desirable for looped Transformers.
However, existing KV quantization methods~\citep{hooper2024kvquant,liu2024kivi}, including state-of-the-art rotation-based approaches~\citep{ashkboos2024quarot,yun2026optr,zhou2026oscar}, often suffer substantial accuracy degradation in sub-4-bit regimes.
Looped Transformers, however, present a unique opportunity: KV states from different loops are closely related because they are generated by repeatedly applying the same shared modules.
This creates strong similarity between KV states across loops as shown in Figure~\ref{fig:e1_loop1_loop2}, which can be exploited to improve low-bit quantization.

Motivated by this \textit{inter-loop similarity}, we propose
\textsc{ResidualQuant}, a loop-aware KV cache quantization method that
represents each loop's KV states as \textit{residuals} from those of a reference
\textit{anchor} loop.
The resulting residuals (e.g., Figure~\ref{fig:e1_loop1_loop2}, right) are more robust to low-bit quantization, enabling
aggressive KV cache compression using INT2 precision while preserving loop-specific information. Our main contributions can be summarized as follows:
\vspace{-3mm}
\begin{itemize}[leftmargin=5mm]
    \item We introduce \textsc{ResidualQuant}, which reconstructs
    loop-specific KV states using the final-loop KV states as an anchor
    and stores their differences as low-precision residuals
    (Section~\ref{sec:reference-selection}).
    We further apply least-squares residual scaling, learned rotations
    (Section~\ref{sec:shared-rotation}), and loop-wise INT4/INT2 mixed precision
    (Section~\ref{sec:loopwise-precision}), which together enable an effective balance between
    low-bit quantization accuracy and runtime efficiency.

    \vspace{1mm}
    \item We show that \textsc{ResidualQuant} substantially improves the
    accuracy--memory tradeoff over state-of-the-art rotation-based KV quantization
    across Ouro-1.4B, Ouro-2.6B, and Huginn-3.5B on reasoning benchmarks.
    In particular, our INT2/4 configuration reduces theoretical KV storage by 80.7\% relative
    to BF16 while retaining accuracy close to BF16, whereas baseline methods degrade sharply
    at the same KV cache budget (Section~\ref{sec:main-results}).

    \vspace{1mm}
    \item We demonstrate that these KV memory savings translate directly into
    higher inference throughput.
    On an RTX 5090, \textsc{ResidualQuant} accelerates decoding by up to
    $3.66\times$ at a fixed batch size. KV compression also supports up to
    $8\times$ larger tested batches and increases peak decode throughput by up to
    $5.38\times$ over BF16 (Section~\ref{sec:throughput-experiments}).

    \end{itemize}

\section{Preliminary}
\label{sec:preliminary}

\subsection{Group-wise Affine Quantization}
Low-bit quantization requires adapting the quantization range to the value distribution.
{To allow this range to vary across channels, the channel dimension of each token and KV head is partitioned into groups $\mathcal G_j$ of size $g$. Each group shares a stored scale $\bar s_j>0$ and offset $\bar o_j$, which determine the spacing and starting point of its quantization levels. For $i\in\mathcal G_j$, integer-coded affine quantization is given by}~\citep{bhalgat2020lsqplus,liu2024kivi,yun2026optr}
\begin{equation}
 \widehat x_i=\bar o_j+\bar s_j\operatorname{clip}_{[0,2^b-1]}\!\left(\operatorname{round}\frac{x_i-\bar o_j}{\bar s_j}\right).
 \label{eq:affine-qdq}
\end{equation}
{Each quantized code uses $b$ bits, and $Q$ denotes quantization followed by reconstruction using these codes and metadata.}

At extreme quantization such as INT2, zero-centered symmetric quantization, where the offset is zero and therefore does not need to be stored, can use only three distinct reconstruction levels, \(\{-\bar s_j,0,\bar s_j\}\), despite having four codes.
Affine quantization allows the reconstruction levels to shift through $\bar o_j$, using all four levels, $\{\bar o_j,\bar o_j+\bar s_j,\bar o_j+2\bar s_j,\bar o_j+3\bar s_j\}$.
Therefore, throughout this work, we adopt affine quantization to fully utilize the INT2 code space while adapting the range to each group's distribution.

\subsection{Rotation-based Quantization}
In \Eqref{eq:affine-qdq}, outliers can increase the group scale $\bar s_j$, widening the spacing between quantization levels and reducing resolution for smaller values. Orthogonal rotations are introduced to redistribute these large magnitudes across channels to mitigate this effect~\citep{chee2023quip,ashkboos2024quarot}. Given an orthogonal matrix $U$, quantization and reconstruction can be expressed as
\begin{equation}
 Q^U(X)=U^\top Q(UX),\qquad \|UX\|_2=\|X\|_2.
 \label{eq:rotation-qdq}
\end{equation}

Existing approaches range from randomized Hadamard rotations~\citep{ashkboos2024quarot} to learned rotations~\citep{liu2025spinquant}. For INT2 KV cache quantization, we choose OptR~\citep{yun2026optr}, which optimizes head-wise orthogonal rotations to minimize attention output error. Specifically, we use its Hadamard-initialized variant, OptR-H, to obtain $U$ for inter-loop residual quantization (Section~\ref{sec:shared-rotation}).

\FloatBarrier
\section{Methodology}
\label{sec:method}

In this section, we present \textsc{ResidualQuant}, our loop-wise residual quantization method for looped Transformers. We first motivate inter-loop residual quantization (Section~\ref{sec:inter-loop-predictability}). Starting from direct INT2 KV cache quantization, we then progressively improve performance by introducing Last-loop residual quantization with least-square (LS) scaling (Section~\ref{sec:residual-formulation}), shared rotations for anchors and residuals (Section~\ref{sec:shared-rotation}), and loop-wise mixed precision (Section~\ref{sec:loopwise-precision}), as illustrated in Figure~\ref{fig:acc-throughput}. Together, these components form \textsc{ResidualQuant}, which recovers BF16-level performance across the evaluated models and benchmarks (Table~\ref{tab:main-benchmarks-reevaluation-method-rows-lowbit}).

\subsection{Motivation}
\label{sec:inter-loop-predictability}
Looped Transformers repeatedly process the same token through shared blocks and apply the same key and value projections across loops. At a fixed layer, token, and head, write $X^{(r)}\in\{K^{(r)},V^{(r)}\}$ as $X^{(r)}=W_Xz^{(r)}$, where $z^{(r)}$ is the preprocessed input and $W_X$ is the projection shared across loops. It follows that
\begin{equation}
 \|X^{(r)}-X^{(r-1)}\|_2
 \leq\|W_X\|_2\,\|z^{(r)}-z^{(r-1)}\|_2.
 \label{eq:state-to-kv}
\end{equation}
Thus, when recurrent inputs change little between loops, the corresponding KV differences are bounded by this change scaled by $\|W_X\|_2$. The fixed-point perspective of deep equilibrium models~\citep{bai2019deq} provides one explanation for small changes across iterations, while Huginn-3.5B exhibits input-dependent trajectories approaching fixed points or forming orbits~\citep{geiping2025huginn}.

Figure~\ref{fig:e1_loop1_loop2} shows that keys exhibit similar patterns across loops, allowing one loop's keys to be represented using another with substantially smaller \textit{inter-loop residuals}. These residuals contain far fewer outliers than the original keys and are therefore easier to quantize. Motivated by the shared-projection structure and this observation, we use a reference KV state to capture shared information and store loop-specific differences as low-precision residuals.

\subsection{Residual Quantization for KV Cache}
\label{sec:residual-formulation}
\label{sec:prediction-coefficient}
\label{sec:reference-selection}
\begin{figure}[t]
 \centering
 \includegraphics[width=\linewidth]{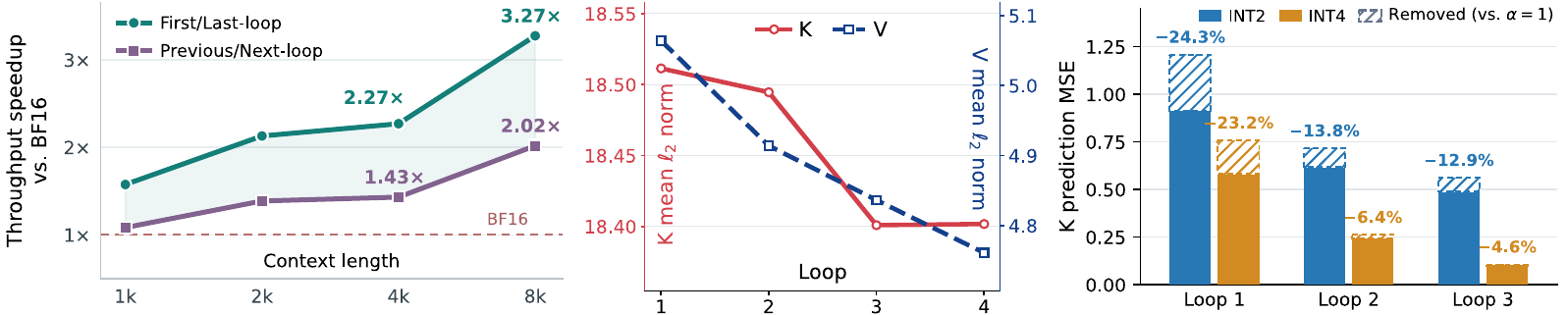}
 \caption{\textbf{Loop-wise KV statistics and the effects of prediction and reference choices on Ouro-1.4B.}
 Left: Decode throughput at maximum feasible batch sizes, with INT4 anchors and INT2 residuals, using group size $g=32$ on an RTX 5090, comparing the First-/Last-loop and Previous-/Next-loop reference families with OptR-H rotations. The plotted measurements use Last-loop and Next-loop, respectively.
Speedups are relative to BF16 vLLM at its maximum feasible batch size.
 Middle: Mean K and V norms across loops.
 Right: Key prediction MSE using Last-loop references under uniform INT2 and INT4, with group size $g=32$ and no rotation.
 Solid bars show prediction MSE with LS scaling, while hatched portions indicate the additional error without LS scaling ($\alpha=1$).
 Analysis details are provided in Appendix~\ref{app:analysis-details}.}
 \label{fig:residual-analysis}
 \label{fig:loop-kv-mean-norm}
\end{figure}

With direct INT2 affine quantization at group size $g=32$, Ouro-1.4B achieves only 27.8\% MATH500 accuracy (Figure~\ref{fig:acc-throughput}).
Motivated by Section~\ref{sec:inter-loop-predictability}, we reconstruct each loop's KV using a reference KV state and quantize only the residual.
For loop $r$, let $X_r\in\mathbb R^D$ denote a key or value vector at a fixed layer, token, and head.
A reference $R_r$ from the stored cache and a scalar coefficient $\alpha_r$ give
\begin{equation}
 \widehat X_r=\alpha_r R_r+Q\!\left(X_r-\alpha_r R_r\right),
 \label{eq:residual-general}
\end{equation}
Here, we store only the quantized residual  $Q\!\left(X_r-\alpha_r R_r\right)$, and reconstruct $\widehat X_r$ on demand using the reference $R_r$.
The most intuitive choice, the \emph{Previous-loop policy},  quantizes the difference from the preceding loop, $X_r-\widehat X_{r-1}$, with $R_r=\widehat X_{r-1}$ and $\alpha_r=1$.
The first loop is quantized separately, i.e., $\widehat X_1=Q(X_1)$, to serve as an anchor for reconstruction.
However, we find this suboptimal and jointly select the reference KV $R_r$ and scaling coefficient $\alpha_r$ to improve model quality while retaining efficient reconstruction.

\begin{wraptable}{r}{0.43\textwidth}
\vspace{-\intextsep}
\centering
\footnotesize
\setlength{\tabcolsep}{2pt}
\captionsetup{font=footnotesize,skip=4pt}
\caption{\textbf{Reference choice on Ouro-1.4B.} Uniform INT2 with group size $g=32$. NMSE compares original and reconstructed keys on all 500 MATH500 problems (Appendix~\ref{app:math500-reference-nmse}).
}
\label{tab:reference-direction-inline}
\begin{tabular}{>{\raggedright\arraybackslash}p{\dimexpr0.24\linewidth-2\tabcolsep\relax}*{5}{>{\raggedleft\arraybackslash}p{\dimexpr0.152\linewidth-2\tabcolsep\relax}}}
\toprule
 & \multicolumn{5}{c}{Key NMSE $\times10^3$ $\downarrow$}\\
\cmidrule(lr){2-6}
Reference & Loop 1 & Loop 2 & Loop 3 & Loop 4 & Avg\\
\midrule
First-loop & 179.39 & 44.88 & 51.71 & 54.30 & 82.57\\
\rowcolor{blue!6}
Last-loop & 53.55 & 28.17 & 19.30 & 178.20 & 69.80\\
\bottomrule
\end{tabular}
\end{wraptable}

\textbf{Reference KV $R_r$.}
The \emph{Previous-loop} policy reconstructs loop~4 in Ouro-1.4B by loading the first-loop anchor and residuals from loops~2--4, requiring four KV cache reads versus one with direct quantization. This increases memory traffic and limits decoding throughput.
An alternative is the \emph{First-loop} policy, which directly references the first-loop anchor, $R_r=\widehat X_1$ for $r=2,\ldots,L$. This allows each non-anchor loop to load only the anchor and its residual.
Figure~\ref{fig:residual-analysis} (left) shows that shared-anchor reconstruction, as used by First-loop, improves decode throughput by up to 1.62$\times$ over chained reconstruction, as used by Previous-loop.

While shared-anchor reconstruction reduces memory traffic, the choice of anchor also affects reconstruction quality. We hypothesize that preserving early-loop KV states is important for maintaining accuracy under aggressive quantization. We therefore propose the \emph{Last-loop} policy, which directly references the final-loop anchor, $R_r=\widehat X_L$ for $r=1,\ldots,L-1$. Table~\ref{tab:reference-direction-inline} compares First-loop and Last-loop under uniform INT2 and shows that Last-loop reduces reconstruction error in early loops, lowering the average reconstruction error across all loops. This reduction can translate into higher accuracy, as suggested by the reference-selection ablation (Table~\ref{tab:math500-reference-scaling} in Section~\ref{sec:ablation}). With Last-loop references under uniform INT2, residual quantization improves Ouro-1.4B's MATH500 accuracy from 27.8\% to 57.2\% (\steptransition{1}{2} in Figure~\ref{fig:acc-throughput}). Detailed reconstruction and access-count comparisons are provided in Appendix~\ref{app:ouro-reconstruction}.

With the \emph{Last-loop} policy, the current token's anchor is unavailable
until the final loop completes.
We therefore retain the current token's KV states in BF16 during decoding
and defer quantization and cache storage until the final loop completes,
as described in Section~\ref{sec:implementation_vllm}.

\textbf{Scaling coefficient $\alpha_r$.}
Figure~\ref{fig:residual-analysis} (middle) shows that KV magnitudes vary across Ouro-1.4B's loops, motivating a scaling coefficient rather than directly subtracting the reference with $\alpha_r=1$. We use the LS coefficient $\alpha_r=\langle X_r,R_r\rangle/\|R_r\|_2^2$ to minimize $\|X_r-\alpha_rR_r\|_2^2$.
Figure~\ref{fig:residual-analysis} (right) shows that, using the same reconstructed reference, LS reduces key prediction MSE measured before residual quantization relative to $\alpha_r=1$.
With Last-loop references and uniform INT2, LS scaling improves MATH500 accuracy from 57.2\% to 60.6\% (\steptransition{2}{3} in Figure~\ref{fig:acc-throughput}).

\subsection{Shared Rotation for Anchors and Residuals}
\label{sec:shared-rotation}
While residual quantization with LS scaling and the Last-loop policy could mitigate KV outliers, some outliers can still remain in the residuals (Figure~\ref{fig:e1_loop1_loop2}, right).
We further reduce their impact using a shared orthogonal rotation $U$ for the anchor and residuals, redistributing outliers across channels before quantization.
We first quantize and reconstruct the final-loop anchor in the rotated space, yielding $\widehat X_L=U^\top Q(UX_L)$. Each non-anchor loop uses this reconstructed anchor as its reference, $R_r=\widehat X_L$, and quantizes its residual with the same rotation $U$. Incorporating shared rotation into \Eqref{eq:residual-general} gives
\begin{equation}
 \widehat X_r=\alpha_r R_r+U^\top Q\!\left(U(X_r-\alpha_r R_r)\right),\; U^\top U=I.
 \label{eq:rotated-residual}
\end{equation}
We use OptR-H~\citep{yun2026optr} to learn separate rotations for keys and values at each layer and head. The anchor and residuals within each loop group share the same rotations.
These fixed matrices are reused across tokens, so their storage does not grow with sequence length.
Applying shared rotations to the INT2 residual quantization obtained in Section~\ref{sec:reference-selection}
improves Ouro-1.4B's MATH500 accuracy from 60.6\% to 71.6\%, with an 11.0\% gain (\steptransition{3}{4} in Figure~\ref{fig:acc-throughput}).

\subsection{Loop-wise Mixed Precision}
\label{sec:loopwise-precision}
To further close the accuracy gap with BF16, we assign INT4 to the final-loop anchor while retaining INT2 for residuals, since the anchor serves as the shared reference for reconstructing all other loops under the \emph{Last-loop} policy.
For Ouro-1.4B with four loops, this yields a precision configuration of $[2,2,2,4]$, increasing the average bitwidth from 2 to 2.5, excluding metadata.
Including quantization metadata, this configuration achieves a 5.2$\times$ reduction in KV memory footprint relative to BF16.
This loop-wise mixed-precision allocation improves MATH500 accuracy
from 71.6\% with uniform INT2 to 76.0\%, matching the BF16 accuracy
with only a 19\% increase in KV cache storage
(\steptransition{4}{5} in Figure~\ref{fig:acc-throughput}).
Combining LS scaling, the Last-loop policy, rotation-based quantization,
and anchor-focused mixed precision, we obtain our proposed method,
\textsc{ResidualQuant}.

\section{Implementation}
\label{sec:implementation}

Accelerating \textsc{ResidualQuant} requires that reconstruction costs do not offset the memory accesses saved by compression. Referencing the final loop also entails handling the current token before its anchor becomes available.
{We address both with a vLLM~\citep{kwon2023efficientmemorymanagementlarge} integration that defers cache storage until the final loop completes and attention kernels that consume compressed KV directly.}

\begin{figure}[!htbp]
\centering
\includegraphics[width=0.70\linewidth]{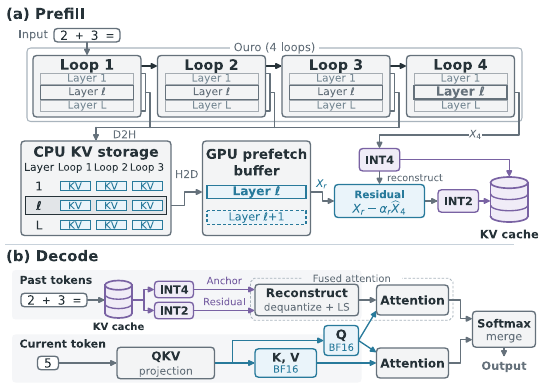}
\caption{\textbf{Execution of \textsc{ResidualQuant} on Ouro.} (a) Prefill: Earlier-loop BF16 KV is offloaded to CPU KV storage and prefetched back into a GPU prefetch buffer per layer for final-loop quantization. (b) Decode: Attention over compressed past KV and the current token's BF16 KV is combined through softmax merge.
}
\label{fig:execution-overview}

\end{figure}

\subsection{vLLM Integration}
\label{sec:implementation_vllm}

We integrate a custom attention backend for compressed KV caches while retaining vLLM's request scheduling and paged KV management.
CUDA kernels jointly quantize and store anchors and residuals in one kernel launch, with separate prefill and decode paths illustrated in Figure~\ref{fig:execution-overview}.

\textbf{Prefill (Figure~\ref{fig:execution-overview}a).}
{Each loop computes causal attention with FlashAttention using its BF16 prompt KV states.}
The Last-loop policy requires retaining non-anchor loops' BF16 prompt KV states until the final-loop anchor becomes available, which can increase peak GPU memory during prefill.
{To reduce this temporary storage, we optionally offload these states to CPU KV storage in pinned host memory.
After each layer in a non-anchor loop computes attention, we copy its BF16 KV into a shared GPU staging buffer, a temporary holding area for asynchronous transfer to CPU KV storage. These device-to-host (D2H) transfers overlap with the layer's output projection and MLP and subsequent-layer computation.}
{Host-to-device (H2D) prefetching for the initial layers begins in the penultimate loop.
During the final loop, a GPU prefetch buffer holds KV states for upcoming layers in a fixed number of cyclically reused slots.
H2D transfers overlap with current-layer computation, while CUDA events prevent slot reuse until transfers and quantization complete.
After each final-loop layer's attention, we await its prefetched KV, quantize the anchor, and use its reconstruction to compute and store the earlier loops' LS-scaled residuals.}
{The full prompt is quantized in one call, and the anchor and residuals are stored as packed codes and metadata in the compressed KV cache.}
Section~\ref{sec:prefill-offloading} evaluates the prefill memory and latency tradeoff.

\textbf{Decode (Figure~\ref{fig:execution-overview}b).}
{During decoding, attention uses compressed KV states of past tokens and BF16 KV states of the current token, combining their contributions through softmax merge. Non-anchor loops retain the current token's BF16 KV in temporary GPU buffers until the final-loop anchor becomes available. After each layer's final-loop attention, we quantize and store the current token's anchor and LS-scaled residuals.}

\subsection{Kernel Implementation}
\label{sec:implementation_kernel}

\textbf{KV quantization and storage.} A fused CUDA kernel quantizes the anchor, reconstructs it to compute the other loops' residuals with LS scaling, and quantizes these residuals. Packed codes and metadata for both are stored in the same kernel launch.

\textbf{Decode attention.} Three CUDA kernels implement decoding attention. For computational efficiency, \textsc{ResidualQuant} shares K and V rotations within groups of four consecutive loops by default, separately for each layer and head.
\textbf{(i) Query preparation and scoring for the current token.}
We compute attention over the current and past tokens separately, as the current token's KV remains in BF16 while past tokens' KV caches are residual-quantized. The prepare kernel computes the current token's attention score using the BF16 query and key.
It also outputs the rotated query, which is used for attention over past tokens and shared between the anchor and residual computations.
\textbf{(ii) Compressed KV attention.}
For past tokens, the \textit{attention kernel} processes their packed anchor and residual codes tile by tile
to compute partition-wise softmax statistics and weighted-V sums. For K, linearity of the dot product allows the kernel to compute scores from the rotated query and unpacked codes with group-wise scale, offset, and LS corrections, without explicitly reconstructing K vectors.
For V, computing the output requires the reconstructed values. Since the anchor and residuals are quantized after rotation, the kernel directly reconstructs each token's value in those rotated coordinates and immediately accumulates its attention-weighted contribution.
This avoids writing reconstructed KV to GPU memory and reading it back. Asynchronous loads overlap the next tile's memory accesses with computation on the current tile. When no explicit V inverse rotation is needed, this kernel incorporates the current token's BF16 contribution and writes the final output for single-partition requests. \textbf{(iii) Output merging.} The \textit{merge} kernel combines partition outputs using their softmax normalization statistics. When the V rotation is shared across all loop groups, it is absorbed into the value and output projection weights at model loading, eliminating explicit V rotation and inverse rotation during attention. For loop-group-specific V rotations, the kernel inverse-rotates the weighted-V sum over past tokens before adding the current token's BF16 contribution.
Softmax statistics and partition merging use FP32 for stability, while QK and weighted-V computations also use BF16 vector arithmetic.

\section{Experiments}
\label{sec:experiments}
Building on the MATH500 improvements in Section~\ref{sec:method}, we evaluate whether \textsc{ResidualQuant} maintains accuracy across multiple benchmarks and experimental settings and translates KV storage savings into higher throughput.
The evaluation begins with accuracy and KV storage comparisons across models (Section~\ref{sec:main-results}), followed by component-wise ablations of \textsc{ResidualQuant} (Section~\ref{sec:ablation}). Throughput measurements assess the gains from KV compression (Section~\ref{sec:throughput-experiments}), and prefill experiments evaluate the memory savings and latency cost of CPU offloading (Section~\ref{sec:prefill-offloading}). Activation quantization experiments test the broader applicability of residual quantization (Section~\ref{sec:activation-extension}).

\afterpage{\begin{table}[!t]
\caption{\textbf{Downstream performance.}
Results are reported for uniform INT2 and mixed INT2/4 with INT2 group sizes
$g\in\{16,32\}$, as indicated in the table.
For mixed INT2/4, INT4 uses group size $g=32$.
Residual quantization uses Last-loop references and LS scaling.
The $b_{\mathrm{eff}}$ value in each group heading reports the effective bits
per KV value for direct quantization and OptR-H.
Avg. denotes the unweighted mean of the twelve model--benchmark scores.
}
\label{tab:main-benchmarks-reevaluation-method-rows-lowbit}
\centering
\scriptsize
\setlength{\tabcolsep}{3.5pt}
\renewcommand{\arraystretch}{1.12}
\begin{adjustbox}{max width=\linewidth}
\begin{tabular}{@{}l*{13}{c}@{}}
\toprule
 & \multicolumn{4}{c}{Ouro-1.4B (4 loops)} & \multicolumn{4}{c}{Ouro-2.6B (4 loops)} & \multicolumn{4}{c}{Huginn-3.5B (32 loops)} & \\
\cmidrule(lr){2-5}\cmidrule(lr){6-9}\cmidrule(lr){10-13}
Method & GSM8K & MATH500 & HumanEval & MBPP & GSM8K & MATH500 & HumanEval & MBPP & GSM8K & MATH500 & HumanEval & MBPP & Avg.\\
\midrule
BF16 & 79.23 & 75.00 & 71.95 & 74.34 & 86.58 & 79.80 & 79.27 & 81.75 & 43.14 & 16.40 & 23.78 & 39.68 & 62.58 \\
\midrule
\multicolumn{14}{l}{\textbf{Uniform INT2}, \textit{INT2 $g=32$} ($b_{\mathrm{eff}}=2.5$)}\\
Direct & 33.66 & 27.80 & 41.46 & 55.56 & 68.54 & 62.40 & 65.85 & 71.43 & 18.95 & 7.80 & 20.73 & 33.60 & 42.32 \\
OptR-H & 62.70 & 55.40 & 59.15 & 70.90 & 81.50 & 72.20 & 72.56 & 73.28 & 28.58 & 10.20 & 22.56 & 36.51 & 53.80 \\
\rowcolor{blue!6}Residual Quantization & 64.67 & 60.60 & 55.49 & 66.40 & 79.53 & 73.60 & \textbf{78.66} & 76.46 & 34.95 & 9.80 & \textbf{24.39} & 36.24 & 55.07 \\
\rowcolor{blue!6}Residual Quantization + OptR-H & \textbf{73.31} & \textbf{71.60} & \textbf{68.90} & \textbf{71.69} & \textbf{83.85} & \textbf{75.20} & 77.44 & \textbf{80.16} & \textbf{38.21} & \textbf{12.60} & 23.78 & \textbf{38.89} & \textbf{59.64} \\
\midrule
\multicolumn{14}{l}{\textbf{Uniform INT2}, \textit{INT2 $g=16$} ($b_{\mathrm{eff}}=3.0$)}\\
Direct & 55.95 & 54.20 & 62.20 & 66.40 & 78.77 & 71.40 & 75.61 & 74.34 & 24.94 & 7.00 & 21.95 & 31.22 & 52.00 \\
OptR-H & 65.96 & 64.20 & \textbf{70.73} & 70.90 & 83.40 & 75.00 & 72.56 & 76.72 & 31.61 & 10.80 & 21.34 & \textbf{40.74} & 57.00 \\
\rowcolor{blue!6}Residual Quantization & 70.96 & 67.80 & 69.51 & 70.37 & 84.15 & 76.60 & \textbf{77.44} & 77.51 & 36.77 & \textbf{14.40} & 25.61 & 37.57 & 59.06 \\
\rowcolor{blue!6}Residual Quantization + OptR-H & \textbf{75.06} & \textbf{75.20} & 70.12 & \textbf{71.16} & \textbf{85.22} & \textbf{79.00} & 75.61 & \textbf{81.22} & \textbf{38.97} & \textbf{14.40} & \textbf{26.22} & 39.42 & \textbf{60.97} \\
\midrule
\multicolumn{14}{l}{\textbf{Mixed INT2/4}, \textit{INT2 $g=32$} ($b_{\mathrm{eff}}=3.0$)}\\
Direct & 48.67 & 47.40 & 56.71 & 67.99 & 76.27 & 73.20 & 70.12 & 78.04 & 30.10 & 9.80 & 21.95 & 34.39 & 51.22 \\
OptR-H & 65.73 & 63.00 & 68.29 & 70.63 & 84.76 & 78.20 & 78.66 & 78.31 & 36.39 & 11.00 & 23.17 & 39.15 & 58.11 \\
\rowcolor{blue!6}Residual Quantization & \textbf{77.79} & 73.80 & \textbf{71.34} & 73.02 & \textbf{86.66} & \textbf{80.20} & 79.88 & 81.75 & 42.46 & \textbf{14.60} & 23.17 & \textbf{40.74} & 62.12 \\
\rowcolor{blue!6}\setbox0=\hbox{Residual quantization + OptR-H}\parbox[c]{\wd0}{\raggedright Residual Quantization + OptR-H} & 77.03 & \textbf{76.00} & 70.73 & \textbf{73.81} & 86.35 & 79.60 & \textbf{80.49} & \textbf{82.28} & \textbf{42.76} & \textbf{14.60} & \textbf{23.78} & 40.21 & \textbf{62.30} \\
\midrule
\multicolumn{14}{l}{\textbf{Mixed INT2/4}, \textit{INT2 $g=16$} ($b_{\mathrm{eff}}=3.4$)}\\
Direct & 63.15 & 61.80 & 66.46 & 72.75 & 81.96 & 76.60 & 73.78 & 79.63 & 32.83 & 11.60 & \textbf{23.17} & 36.24 & 56.66 \\
OptR-H & 69.98 & 67.00 & \textbf{73.78} & 71.16 & 86.05 & 78.60 & 77.44 & 80.16 & 36.24 & 10.20 & 21.34 & 37.83 & 59.15 \\
\rowcolor{blue!6}Residual Quantization & 77.03 & \textbf{75.00} & 71.95 & \textbf{74.07} & 85.82 & \textbf{80.40} & \textbf{79.27} & 80.95 & 41.02 & 15.40 & \textbf{23.17} & \textbf{41.53} & 62.13 \\
\rowcolor{blue!6}\setbox0=\hbox{Residual Quantization + OptR-H}\parbox[c]{\wd0}{\raggedright Residual Quantization + OptR-H} & \textbf{77.63} & 74.60 & 72.56 & 73.28 & \textbf{87.41} & \textbf{80.40} & \textbf{79.27} & \textbf{82.01} & \textbf{42.61} & \textbf{15.60} & 22.56 & 39.42 & \textbf{62.28} \\
\bottomrule
\end{tabular}
\end{adjustbox}
\end{table}
}

\paragraph{Models.}
We evaluate our residual quantization method on Ouro-1.4B, Ouro-2.6B, and Huginn-3.5B. Ouro models execute four loops, while Huginn-3.5B executes 32 loops. For Huginn-3.5B, we partition the 32 loops into eight groups of four and apply \textsc{ResidualQuant} independently within each group, using the final loop of each group as its anchor.

\paragraph{Quantization setup.}
For KV quantization experiments, model weights remain in BF16, while KV caches use BF16, uniform INT4, uniform INT2, or loop-wise mixed INT2/4, with default group sizes $g=16$ for uniform INT2 and $g=32$ for INT4 and mixed INT2/4 unless otherwise specified. The mixed configuration assigns INT4 to the anchor loop and INT2 to the remaining loops (Section~\ref{sec:loopwise-precision}).

For a fair comparison, all quantized methods use the same KV handling during prefill and decoding.
During prefill, each loop computes attention using BF16 prompt KV states before quantizing and storing them for decoding.
During decoding, the current token's KV states remain in BF16, as required by
our Last-loop implementation, and we apply the same treatment to all baseline
methods.
As baselines, we consider direct uniform quantization and the rotation-based OptR-H method~\citep{yun2026optr}. Rotation calibration and inference details are provided in Appendix~\ref{app:rotation-details}.

\paragraph{Benchmarks.}
For mathematical reasoning, we report accuracy on GSM8K~\citep{cobbe2021gsm8k} and MATH500 \citep{lightman2024verify}. For code generation, we report pass@1 on HumanEval~\citep{chen2021humaneval} and MBPP~\citep{austin2021mbpp}. Prompt formats, generation limits, and evaluation details are provided in Appendix~\ref{subsec:details-benchmarks-and-prompts}.

\subsection{Main Results}
\label{sec:main-results}
We compare direct uniform quantization, OptR-H, and residual quantization
with Last-loop references and LS scaling, evaluated both with and without
rotation, on Ouro-1.4B, Ouro-2.6B, and Huginn-3.5B.
Evaluating residual quantization without rotation isolates its own contribution. The mixed INT2/4 configuration with rotation constitutes our full \textsc{ResidualQuant} method. We denote the corresponding uniform INT2 variant as \textsc{ResidualQuant-INT2}. For each method, we evaluate uniform INT2 and mixed-precision INT2/4 with INT2 group sizes $g=16$ and $g=32$ to characterize the accuracy--memory tradeoff.

The main evaluation examines the accuracy--memory tradeoff using the effective
bitwidth $b_{\mathrm{eff}}$, which measures the total KV cache storage cost per
KV element, including quantized values and associated metadata.
The $b_{\mathrm{eff}}$ values in
Table~\ref{tab:main-benchmarks-reevaluation-method-rows-lowbit} report the
storage cost for direct quantization and OptR-H.
Residual quantization additionally stores BF16 LS coefficients, increasing
$b_{\mathrm{eff}}$ by approximately $0.1$ bit.
For example, under uniform INT2 with $g=32$, direct quantization and OptR-H
use $2.5$ bits per KV element, whereas residual quantization uses approximately
$2.6$ bits.
The full storage accounting is provided in
Appendix~\ref{subsec:effective-kv-bitwidth}.

Table~\ref{tab:main-benchmarks-reevaluation-method-rows-lowbit} reports results
across three models and four benchmarks. The Avg. column is the unweighted
mean of these twelve scores.
Across all four precision configurations, residual quantization without
rotation achieves a higher average score than the rotation-based OptR-H
baseline at comparable KV budgets.
The improvement reaches 4.01\% under mixed INT2/4 with
$g=32$ (62.12 vs.\ 58.11), supporting the use of inter-loop residual
structure for low-bit KV quantization in looped Transformers.

OptR-H can further be applied orthogonally to residual quantization,
providing complementary gains across most evaluated settings.
Under mixed INT2/4 with $g=32$, the full \textsc{ResidualQuant} method
nearly matches the BF16 average score (62.30 vs.\ 62.58) while reducing
theoretical KV storage by 80.7\%.
At the same precision configuration, direct quantization and OptR-H fall
11.36 and 4.47\% below BF16, respectively.
Under the more aggressive uniform INT2 setting with $g=32$,
\textsc{ResidualQuant-INT2} is 2.94\% below BF16 on average, compared
with gaps of 20.26 and 8.78\% for direct quantization and OptR-H.
For example, on Ouro-1.4B MATH500, \textsc{ResidualQuant-INT2} achieves
71.60\% compared with 55.40\% for OptR-H, a 16.20\% improvement.

\begin{table}[!t]
\caption{\textbf{LS scaling, rotation, and reference selection ablation on Ouro-1.4B.}
MATH500 accuracy (\%). Uniform INT2 uses group size $g=16$, while mixed INT2/4 uses $g=32$ in every loop.
For mixed INT2/4, Previous-loop and First-loop use the $[4,2,2,2]$
configuration, while Next-loop and Last-loop use $[2,2,2,4]$, assigning
the higher precision to the anchor loop in each case.
The w/o and w/ LS scaling columns report residual quantization without rotation. The w/ LS scaling + OptR-H column reports residual quantization with LS scaling and OptR-H.}
\label{tab:math500-reference-scaling}
\centering
\footnotesize
\setlength{\tabcolsep}{4pt}
\begin{adjustbox}{max width=\linewidth}
\begin{tabular}{l*{6}{c}}
\toprule
& \multicolumn{3}{c}{INT2}
& \multicolumn{3}{c}{INT2/4} \\
\cmidrule(lr){2-4}\cmidrule(lr){5-7}
Reference
& w/o LS scaling & w/ LS scaling & w/ LS scaling + OptR-H
& w/o LS scaling & w/ LS scaling & w/ LS scaling + OptR-H \\
\midrule

Previous-loop
& 69.2
& 71.2
& 72.2
& 66.2
& 73.2
& 72.8
\\

Next-loop
& 71.6
& 73.2
& 72.8
& 71.4
& 74.0
& 75.2
\\

First-loop
& 46.0
& 69.8
& 71.4
& 24.8
& 73.8
& 72.2
\\

\rowcolor{blue!6}
Last-loop
& 69.2
& 67.8
& 75.2
& 70.2
& 73.8
& 76.0
\\

\bottomrule
\end{tabular}
\end{adjustbox}
\end{table}

\subsection{Ablation Studies}
\label{sec:ablation}
We further evaluate the contributions of the individual components introduced in Section~\ref{sec:method} through ablations of \textsc{ResidualQuant}, including LS scaling, reference choice, rotation, and loop-wise precision allocation.
All accuracy comparisons use Ouro-1.4B on MATH500. A full table comparing multiple scaling methods, rotations, and reference choices is provided in Appendix~\ref{app:full-kv-ablations}.

\noindent\textbf{LS scaling.}
Table~\ref{tab:math500-reference-scaling} compares residual quantization with and without LS scaling across four reference selection policies and two precision settings (INT2 and INT2/4).
LS scaling improves accuracy in most configurations, demonstrating its effectiveness across different reference choices and precisions.
In particular, under the INT2/4 Last-loop configuration, LS scaling improves MATH500 accuracy from 70.2\% to 73.8\%, a 3.6\% gain.
This improvement is also supported by Figure~\ref{fig:residual-analysis} (right), where LS scaling effectively reduces key prediction MSE by up to 24.3\% relative to the unscaled case.

\paragraph{Reference choice.}
In Table~\ref{tab:math500-reference-scaling}, we compare the Previous-loop, First-loop, and Last-loop policies introduced in Section~\ref{sec:reference-selection}. We additionally evaluate Next-loop, which references the immediately following loop and uses the final loop as its anchor, i.e., the reverse of Previous-loop. Overall, the Last-loop policy achieves accuracy comparable to or better than Previous-loop, while avoiding its chained reconstruction overhead. In particular, with LS scaling and OptR-H, Last-loop improves MATH500 accuracy from 72.2\% to 75.2\% under INT2 and from 72.8\% to 76.0\% under INT2/4 compared with Previous-loop. Together with additional $1.62\times$ throughput improvement from shared-anchor reconstruction shown in Figure~\ref{fig:residual-analysis} (left), these results motivate our choice of Last-loop.

First-loop also avoids chained reconstruction and can provide similar
runtime benefits.
However, it yields substantially lower accuracy than Last-loop: 71.4\% versus 75.2\% under INT2 and 72.2\% versus 76.0\%
under INT2/4.
Next-loop, in contrast, achieves accuracy comparable to Last-loop, but retains
the same chained reconstruction dependency as Previous-loop and therefore
suffers from the same memory-traffic overhead.
We therefore adopt Last-loop as the default reference policy, as it provides the
best balance between quantization accuracy and decoding efficiency.

\begin{wraptable}{r}{0.56\textwidth}
\vspace{-\intextsep}
\centering
\captionsetup{font=footnotesize,skip=4pt}
\setlength{\tabcolsep}{1.5pt}

\caption{\textbf{Residual quantization and rotation.}
MATH500 accuracy (\%) on Ouro-1.4B, with group sizes $g=16$ for INT2 and $g=32$ for INT4.
Residual uses Last-loop references and LS scaling.
OSCAR~\citep{zhou2026oscar} and OptR-H~\citep{yun2026optr} are applied to both quantizers. BF16: 75.0\%.}
\footnotesize
\begin{tabularx}{\linewidth}{@{}l*{6}{>{\centering\arraybackslash}X}@{}}
\toprule
\multirow{2}{*}{Method} & \multicolumn{3}{c}{INT2} & \multicolumn{3}{c}{INT4} \\
\cmidrule(lr){2-4}\cmidrule(lr){5-7}
& None & OSCAR & OptR-H & None & OSCAR & OptR-H \\
\midrule
Direct & 54.2 & 65.8 & 64.2 & 74.2 & 75.0 & 75.2 \\
Residual & 67.8 & 74.8 & 75.2 & 74.2 & 77.2 & 74.4 \\
\bottomrule
\end{tabularx}
\label{tab:math500-rotation-residual}
\end{wraptable}

\paragraph{Rotation.}
To examine whether rotation further improves accuracy, Table~\ref{tab:math500-rotation-residual} compares direct and residual quantization (with LS scaling and the Last-loop policy) with and without rotation.
We evaluate INT2 with group size $g=16$ and INT4 with $g=32$ using two recent rotation methods, OSCAR~\citep{zhou2026oscar} and OptR-H~\citep{yun2026optr}. At INT2, OSCAR and OptR-H improve direct from 54.2\% to 65.8\% and 64.2\%, respectively. Residual quantization achieves 67.8\% without rotation, and combining it with OSCAR or OptR-H further improves accuracy to 74.8\% and 75.2\%, respectively. At INT4, residual quantization achieves 74.2\% without rotation, while OSCAR and OptR-H improve it to 77.2\% and 74.4\%, respectively. These results demonstrate the complementary benefits of residual quantization and rotation, although the gain depends on precision and rotation method.

\begin{table}[!t]
\centering

\caption{\textbf{Loop-wise precision allocation ablations.}
    MATH500 accuracy (\%) on Ouro-1.4B and effective KV bitwidth $b_{\mathrm{eff}}$.
    We compare direct quantization and residual quantization with and without OptR-H rotation.
    Residual quantization uses LS scaling and the Last-loop policy.
    \emph{Uniform} applies the same precision to all loops, \emph{Descending} uses $[4,2,2,2]$, and \emph{Ascending} uses $[2,2,2,4]$, allocating the higher precision to the first and last loop, respectively.
    Group size $g$ shown in the table applies only to INT2, while INT4 uses group size 32.
    Blue shading marks our final INT2/4 precision configuration.
    BF16 accuracy is 75.0\%.
}
\label{tab:math500-segments}

\footnotesize
\setlength{\tabcolsep}{4pt}
\renewcommand{\arraystretch}{1.05}
\begin{tabularx}{\linewidth}{llc*{6}{>{\centering\arraybackslash}X}}
\toprule
\multirow{2}{*}{Schedule}
& \multirow{2}{*}{Config}
& \multirow{2}{*}{$g$}
& \multicolumn{3}{c}{Direct Quantization}
& \multicolumn{3}{c}{Residual Quantization} \\
\cmidrule(lr){4-6}\cmidrule(lr){7-9}
& & & $b_{\mathrm{eff}}$ & w/o Rotation & w/ Rotation
    & $b_{\mathrm{eff}}$ & w/o Rotation & w/ Rotation \\
\midrule
\multirow{3}{*}{\textit{Uniform}}
& $[4,4,4,4]$ & -- & 4.5 & 74.20 & 75.20 & 4.6 & 74.20 & 74.40 \\
& $[2,2,2,2]$ & 32 & 2.5 & 27.80 & 55.40 & 2.6 & 60.60 & 71.60 \\
& $[2,2,2,2]$ & 16 & 3.0 & 54.20 & 64.20 & 3.1 & 67.80 & 75.20 \\
\midrule
\multirow{2}{*}{\textit{Descending}}
& $[4,2,2,2]$ & 32 & 3.0 & 40.20 & 61.80 & 3.1 & 61.80 & 69.00 \\
& $[4,2,2,2]$ & 16 & 3.4 & 57.60 & 63.60 & 3.5 & 71.20 & 74.40 \\
\midrule
\rowcolor{blue!6}
& $[2,2,2,4]$ & 32 & 3.0 & 47.40 & 63.00 & 3.1 & 73.80 & 76.00 \\
\rowcolor{blue!6}
\multirow{-2}{*}{\textit{Ascending}}
& $[2,2,2,4]$ & 16 & 3.4 & 61.80 & 67.00 & 3.5 & 75.00 & 74.60 \\
\bottomrule
\end{tabularx}
\end{table}

\begin{figure}[!t]
 \centering
 \includegraphics[width=0.9\linewidth]{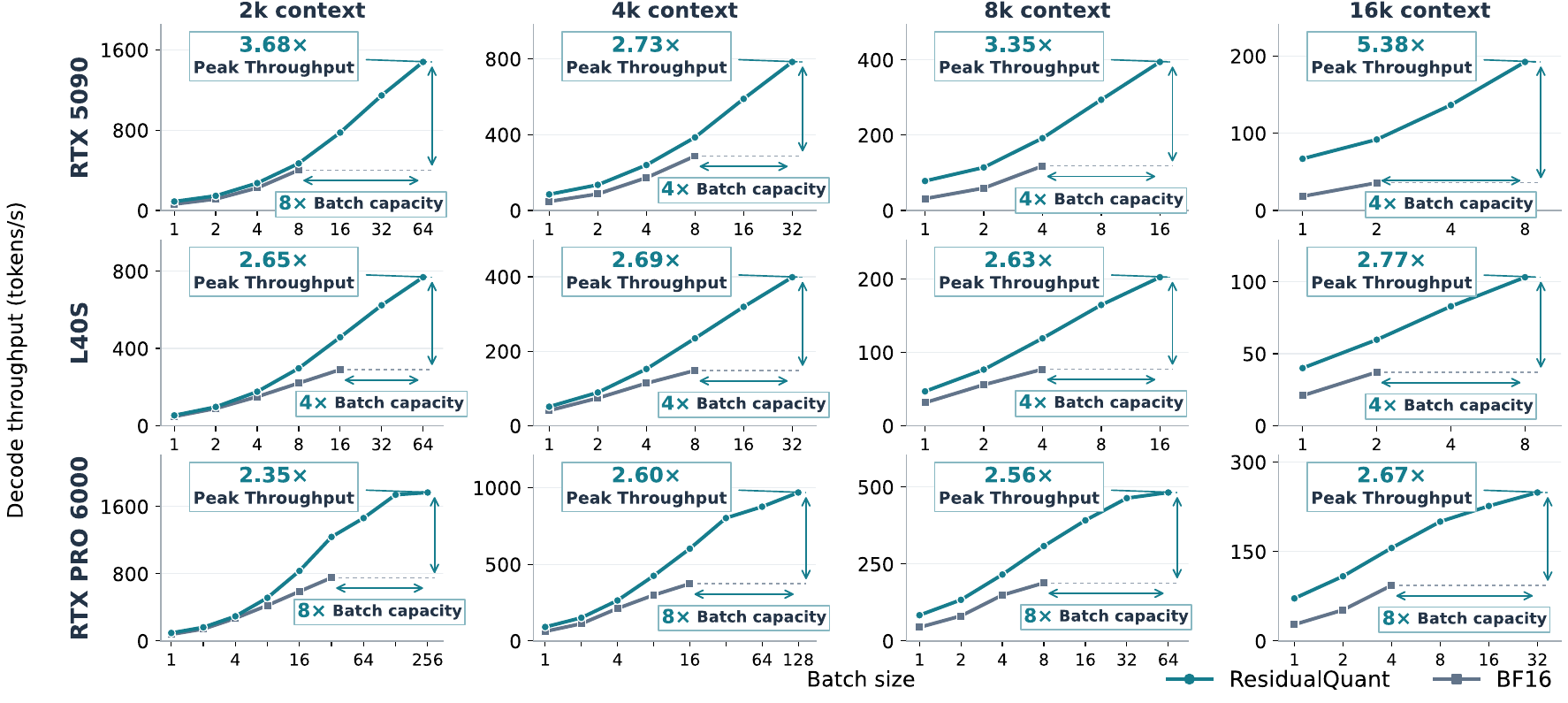}
 \caption{\textbf{Decode throughput across batch sizes and context lengths.} \textsc{ResidualQuant} with group size $g=32$ in every loop and BF16 on Ouro-1.4B on RTX 5090, L40S, and RTX PRO 6000 Blackwell GPUs (top to bottom), at 2k, 4k, 8k, and 16k context (left to right). Vertical and horizontal arrows indicate gains in peak throughput and the largest feasible tested power-of-two batch, respectively.}
 \label{fig:batch-throughput}
\end{figure}

\paragraph{Loop-wise precision allocation.}
Finally, we examine how accuracy depends on loop-wise precision allocation.
Table~\ref{tab:math500-segments} compares uniform precision with \emph{Descending} $[4,2,2,2]$ and \emph{Ascending} $[2,2,2,4]$ schedules.
Under Last-loop residual quantization, Ascending allocates the higher precision to the shared anchor, whereas Descending allocates it to a residual loop.
Ascending outperforms Descending with residual quantization at both group sizes, with and without rotation.
In particular, at $g=32$, Ascending achieves 73.8\% versus 61.8\% without rotation and 76.0\% versus 69.0\% with rotation, supporting the benefit of allocating higher precision to the shared anchor.
Moreover, at the same effective bitwidth ($b_{\mathrm{eff}}=3.1$), Ascending with $g=32$ outperforms uniform INT2 with $g=16$ by 6.0\% and 0.8\% without and with rotation, demonstrating a better accuracy--memory trade-off.

\subsection{Throughput Improvement}
\label{sec:throughput-experiments}
\label{sec:additional-analysis}

KV compression can improve throughput not only by accelerating a fixed batch but also by allowing more requests to run together on one GPU. Figure~\ref{fig:batch-throughput} reports decode throughput at 2k, 4k, 8k, and 16k contexts. The comparison uses Ouro-1.4B on RTX 5090, L40S, and RTX PRO 6000 Blackwell GPUs with \textsc{ResidualQuant} (group size $g=32$ in every loop) and BF16 vLLM FlashAttention, doubling batch sizes from 1 to 256 and plotting measured feasible configurations. Each run generates 128 tokens.

On RTX 5090, at fixed batch sizes, \textsc{ResidualQuant} improves decode throughput at all context lengths. At 8k context and batch size 4, throughput increases from 117.9 to 191.9 tokens/s ($1.63\times$). At 16k context and batch size 1, throughput increases from 18.3 to 66.9 tokens/s ($3.66\times$). These gains are consistent with the reduced KV cache traffic enabled by compression.

\begin{wrapfigure}{r}{0.45\textwidth}
\vspace{-\intextsep}
 \centering
 \captionsetup{font=footnotesize,skip=4pt}
 \includegraphics[width=\linewidth]{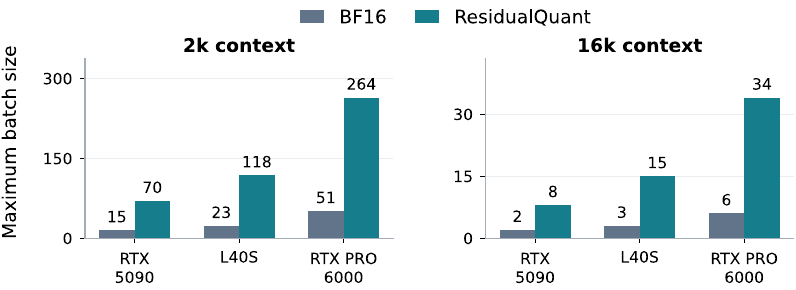}
 \caption{\textbf{Max batch size.} BF16 and \textsc{ResidualQuant} on Ouro-1.4B on RTX 5090, L40S, and RTX PRO 6000. Panels show 2k and 16k contexts. Additional input lengths are reported in Appendix~\ref{app:additional-runtime}.}
 \label{fig:max-batch-capacity}
\end{wrapfigure}

The smaller KV footprint additionally allows larger batches. Figure~\ref{fig:max-batch-capacity} shows that \textsc{ResidualQuant} supports up to $5.67\times$ as many concurrent requests as BF16 at 2k and 16k contexts. At 2k, RTX 5090 supports 15 requests with BF16 and 70 with \textsc{ResidualQuant} ($4.67\times$), while L40S supports 23 and 118 ($5.13\times$), and RTX PRO 6000 supports 51 and 264 ($5.18\times$), respectively. Across all three GPUs and tested context lengths, \textsc{ResidualQuant} increases the largest feasible tested power-of-two batch by $4$--$8\times$ and peak decode throughput by $2.35$--$5.38\times$ over BF16. For example, on RTX 5090 at 2k, the largest feasible tested batch increases from 8 for BF16 to 64 for \textsc{ResidualQuant}, raising throughput to 1479.0 tokens/s ($3.68\times$). At 16k, the corresponding batch increases from 2 to 8, raising throughput to 192.2 tokens/s ($5.38\times$). Thus, faster decoding at a fixed batch and increased batch capacity jointly improve peak throughput. Appendix~\ref{app:additional-runtime} provides the 4k and 8k results.

\subsection{Prefill Memory and Offloading Overhead}
\label{sec:prefill-offloading}

\begin{wraptable}[14]{r}{0.53\textwidth}
\vspace{-\intextsep}
\centering
\captionsetup{font=footnotesize,skip=4pt}
\caption{\textbf{Prefill latency and GPU memory.}
Ouro-1.4B, batch size 1, RTX 5090. None, Serial, and Overlap use full \textsc{ResidualQuant} with group size $g=32$.
Medians of 30 measurements. Latency is in ms and peak GPU memory in GiB. Parentheses report ratios relative to BF16 at each input length.
Additional input lengths are reported in Appendix~\ref{app:additional-runtime}. 
Serial uses a single GPU layer buffer, whereas Overlap uses GPU staging and prefetch buffers.
}
\label{tab:prefill-offloading-main}
\footnotesize
\setlength{\tabcolsep}{1.5pt}
\renewcommand{\arraystretch}{1.05}
\begin{adjustbox}{max width=\linewidth}
\begin{tabular}{@{}*{5}{c}@{}}
\toprule
& \multicolumn{2}{c}{2K} & \multicolumn{2}{c}{16K} \\
\cmidrule(lr){2-3}\cmidrule(lr){4-5}
Method & Latency & Peak & Latency & Peak \\
\midrule
BF16 & 133.4\,{($1.00\times$)} & 4.37\,{($1.00\times$)} & 1426.8\,{($1.00\times$)} & 15.65\,{($1.00\times$)} \\
None & 137.2\,{($1.03\times$)} & 4.26\,{($0.98\times$)} & 1427.4\,{($1.00\times$)} & 14.97\,{($0.96\times$)} \\
Serial & 186.1\,{($1.40\times$)} & 3.18\,{($0.73\times$)} & 1755.6\,{($1.23\times$)} & 6.34\,{($0.41\times$)} \\
\textbf{Overlap} & \textbf{137.4}\,{($\boldsymbol{1.03\times}$)} & \textbf{3.55}\,{($\boldsymbol{0.81\times}$)} & \textbf{1428.9}\,{($\boldsymbol{1.00\times}$)} & \textbf{6.87}\,{($\boldsymbol{0.44\times}$)} \\
\bottomrule
\end{tabular}
\end{adjustbox}

\end{wraptable}

With Last-loop references, KV states from {non-anchor loops} must be retained in BF16 until the final-loop anchor becomes available at each layer, after which the anchor and residuals are quantized and stored.
We evaluate the resulting prefill latency and peak GPU memory relative to BF16 on Ouro-1.4B, using \textsc{ResidualQuant} with group size $g=32$ and batch size 1 on an RTX 5090 with a PCIe 5.0 $\times$16 connection at 2k--16k input lengths.
We compare {no offloading (None)}, serialized full-layer KV offloading (Serial), and full-layer KV offloading with {asynchronous D2H transfers and H2D prefetching} (Overlap).
Serial performs D2H/H2D transfers without overlapping computation and reuses a single GPU layer buffer for H2D transfers. Overlap uses the shared GPU staging buffer and prefetch slots described in Section~\ref{sec:implementation_vllm} to overlap D2H/H2D transfers with computation.
{Both transfer full-layer BF16 KV states and quantize each layer's full prompt in one call after its final-loop attention.}
We report medians of 30 eager prefill forwards after five warm-up requests.
Latency includes D2H/H2D transfers, KV quantization, and storage in the compressed KV cache but excludes decoding.
Peak allocated GPU memory includes model weights, the KV pool, graph buffers, and prefill temporaries, including offloading buffers.

Compared with the historical BF16 baseline in Table~\ref{tab:prefill-offloading-main}, \textsc{ResidualQuant} without offloading increases median prefill latency by at most 3.2\% across 2k--16k inputs, indicating small overall prefill overhead despite KV quantization.

\begin{table}[!t]

\caption{\textbf{Extension of residual quantization to W4A4 activation quantization.}
Accuracy or pass@1 (\%) when applying residual quantization with LS scaling to W4A4 activation quantization, with KV caches kept in BF16.
We consider three baselines: Naive round-to-nearest quantization, FlatQuant~\citep{sun2025flatquant}, and LoopQ~\citep{fang2026loopq}.
Avg. denotes the unweighted mean of the eight model--benchmark scores.
Details are provided in Appendix~\ref{app:activation-details}.}
\label{tab:activation-extension}

\centering
\footnotesize
\setlength{\tabcolsep}{1pt}
\renewcommand{\arraystretch}{1.12}
\begin{tabularx}{\linewidth}{l*{8}{>{\centering\arraybackslash}X}c}
\toprule
& \multicolumn{4}{c}{Ouro-1.4B (4 loops)}
& \multicolumn{4}{c}{Huginn-3.5B (32 loops)} & \\
\cmidrule(lr){2-5}\cmidrule(lr){6-9}
Method & GSM8K & MATH500 & HumanEval & MBPP
       & GSM8K & MATH500 & HumanEval & MBPP & Avg. \\
\midrule
BF16
& 78.77
& 75.00
& 71.95
& 74.87
& 42.46
& 12.40
& 23.17
& 41.01
& 52.45
\\
\midrule
Naive
& 49.51
& 53.00
& 62.20
& 65.08
& 33.36
& 9.40
& 20.12
& 36.77
& 41.18
\\
\rowcolor{blue!6}
\quad + Residual Quantization
& 56.86
& 61.00
& 62.20
& 68.52
& 39.12
& \textbf{12.80}
& 23.17
& \textbf{39.68}
& 45.42
\\
\addlinespace[2pt]
FlatQuant
& 68.31
& 67.40
& 67.68
& 69.58
& 33.21
& 9.20
& 18.29
& 39.15
& 46.60
\\
\rowcolor{blue!6}
\quad + Residual Quantization
& \textbf{70.20}
& \textbf{68.80}
& \textbf{70.12}
& \textbf{71.69}
& 38.06
& 12.60
& \textbf{28.05}
& 37.57
& \textbf{49.64}
\\
\addlinespace[2pt]
LoopQ
& 65.88
& 64.40
& 64.63
& 66.14
& 35.03
& 11.20
& 18.90
& 38.89
& 45.63
\\
\rowcolor{blue!6}
\quad + Residual Quantization
& 68.31
& 64.80
& 67.68
& 70.90
& \textbf{41.02}
& 12.60
& 26.22
& 38.89
& 48.80
\\
\bottomrule
\end{tabularx}
\end{table}

With Overlap, latency remains similar to BF16 (within 3.5\%), indicating that {D2H/H2D transfers} add little observable latency. With this setting, we reduce peak GPU memory by up to 56.1\%, from 15.65 to 6.87 GiB at 16k, by offloading non-anchor loops' BF16 KV states to CPU KV storage. Serial further reduces peak memory to 6.34 GiB, 59.5\% below BF16, because it uses a single GPU layer buffer without GPU staging or prefetch buffers. However, it incurs 23.0--39.5\% higher latency than BF16 because D2H/H2D transfers are serialized with computation.

\subsection{Extension to Activation Quantization}
\label{sec:activation-extension}
The preceding experiments show that inter-loop similarity can be exploited to compress KV states through low-bit residual quantization.
Here, we investigate whether the same principle extends beyond KV storage to activation quantization, which enables low-precision computation in projection layers.
Because looped Transformers repeatedly apply the same projections across loops, activations from one loop can also provide a useful reference for the corresponding projection inputs in another loop.
We therefore incorporate residual quantization into three W4A4 baselines: Naive, FlatQuant~\citep{sun2025flatquant}, an established weight--activation quantization method, and LoopQ~\citep{fang2026loopq}, which is specifically designed for looped Transformers.
Naive applies round-to-nearest quantization to weights and activations with group size $g=32$ without calibration.
Implementation and calibration details for FlatQuant and LoopQ are provided in Appendix~\ref{app:activation-details}.

Unlike baselines that quantize each loop's full activation independently, we quantize $X_r-\alpha_r\widehat X_{r-1}$ using Previous-loop with LS scaling and apply the low-precision projection to this residual. Here, $X_r$ is the current activation and $\widehat X_{r-1}$ is the previous loop's reconstructed activation. The first loop is quantized directly. Previous-loop is used because activations are consumed immediately within each loop. To isolate the effect of residual quantization on activations, weights and activations remain INT4 and KV caches BF16.

Table~\ref{tab:activation-extension} shows that applying residual quantization with LS scaling improves Ouro-1.4B's GSM8K accuracy from 49.51\% to 56.86\% for Naive, from 68.31\% to 70.20\% for FlatQuant, and from 65.88\% to 68.31\% for LoopQ. On Huginn-3.5B, Naive and FlatQuant improve by 5.76 and 4.85\%, respectively. Accuracy improves in nearly all evaluated method--benchmark combinations across both models.
These results show that inter-loop residual quantization extends beyond KV compression and can consistently improve existing low-bit activation quantization methods as well.
\par

\section{Related Work}
\label{sec:related-work}

\paragraph{Optimizing Looped Transformers.}
Parameter sharing does not eliminate the KV storage and sequential execution costs of additional loops. LoopSpec~\citep{cho2026loopspec} accelerates decoding by drafting tokens from early recurrent states and overlapping draft generation with target verification through pipelined self-speculative decoding. MoR selectively caches active tokens and offers first-recursion KV reuse~\citep{bae2025mor}. PLT combines first-loop KV sharing with gated sliding-window attention and inter-loop parallelism~\citep{wu2025plt}. MELT makes cache size independent of loop depth by updating one shared cache per layer through learned gates~\citep{vendrell2026melt}. Further background on looped Transformers is provided in Appendix~\ref{app:detailed-related-work}.

\paragraph{Quantization.}
Quantization offers a complementary direction. LoopQ addresses inter-loop distribution shifts and error accumulation in W4A4 post-training quantization~\citep{fang2026loopq}. For KV compression, KIVI and KVQuant tailor quantization to key and value distributions~\citep{liu2024kivi,hooper2024kvquant}, while rotation-based methods redistribute outliers across channels~\citep{ashkboos2024quarot,liu2025spinquant,su2025rotatekv}. Concurrent FlashLoop~\citep{yang2026flashloop} combines lazy KV updates and quantization motivated by a similar inter-loop observation. A detailed comparison is provided in Appendix~\ref{app:flashloop-comparison}.

\section{Conclusion}
\label{sec:conclusion}
We presented \textsc{ResidualQuant}, a loop-aware KV cache quantization method
that exploits inter-loop similarity by reconstructing loop-specific KV states
from the final-loop KV states and low-precision residuals.
Combining Last-loop references, least-square scaling, rotation-based quantization, and loop-wise mixed precision enables accurate quantization
down to INT2 while retaining efficient reconstruction.

Across multiple models and benchmarks spanning mathematical reasoning and code
generation, \textsc{ResidualQuant} retains accuracy close to BF16 under mixed-precision settings while reducing
theoretical KV storage by up to 80.7\%, and consistently outperforms rotation-based quantization
at comparable memory budgets.
These compression gains also translate into practical system benefits:
on an RTX 5090, \textsc{ResidualQuant} accelerates fixed-batch decoding by up to
$3.66\times$ and, by enabling larger batches, increases peak throughput by up to
$5.38\times$.

\clearpage
\bibliographystyle{plainnat}
\bibliography{paper}

\clearpage
\appendix
\section{Experimental Setup Details}
\label{app:experimental-setup}

\subsection{Models and Hardware}
\label{subsec:details-models-and-hardware}
Ouro-1.4B uses the \texttt{ByteDance/Ouro-1.4B} checkpoint, Ouro-2.6B uses the \texttt{ByteDance/Ouro-2.6B} checkpoint, and Huginn-3.5B uses \texttt{tomg-group-umd/huginn-0125}. Experiments are conducted on NVIDIA RTX 5090, NVIDIA L40S and NVIDIA RTX PRO 6000 Blackwell GPUs. The L40S uses PCIe 4.0 $\times$16, while the RTX 5090 and RTX PRO 6000 Blackwell use PCIe 5.0 $\times$16. Each evaluation uses a single GPU, BF16 weights, a compressed KV backend integrated into vLLM, and decode CUDA Graphs. Both Ouro models execute 4 loops and Huginn-3.5B 32 loops, with residual references confined to each group of four loops.

\subsection{Benchmarks and Prompts}
\label{subsec:details-benchmarks-and-prompts}

\begin{table}[H]
\caption{Prompt formats and maximum generated tokens. Ouro-1.4B uses plain-text prompts, while Huginn-3.5B uses its tokenizer's chat template.}
\label{tab:prompt-settings}
\centering
\small
\begin{tabular}{llcl r}
\toprule
Model & Benchmark & Shots & Prompt format & Max tokens\\
\midrule
\multirow{4}{*}{Ouro-1.4B} & GSM8K & 3 & GSM8K CoT & 1,024\\
 & MATH500 & 0 & CoT, boxed answer & 2,048\\
 & HumanEval & 0 & EvalPlus-style & 1,024\\
 & MBPP & 0 & EvalPlus-style & 2,048\\
\midrule
\multirow{4}{*}{Huginn-3.5B} & GSM8K & 8 & GSM8K CoT & 1,024\\
 & MATH500 & 4 & Adapted Minerva & 2,048\\
 & HumanEval & 0 & EvalPlus-style & 1,024\\
 & MBPP & 0 & EvalPlus-style & 2,048\\
\bottomrule
\end{tabular}
\end{table}

The evaluations cover 1,319 GSM8K problems, 500 MATH500 problems, 164 HumanEval problems, and 378 MBPP problems. For mathematical reasoning, GSM8K is scored by exact match on extracted final answers, and MATH500 is scored using Math-Verify. Code evaluation uses the curated tasks and evaluation tools from EvalPlus \citep{liu2023evalplus}, while the reported metric is base pass@1 on the original tests. All methods use the same model-specific prompts and stopping conditions, generating one response per problem with greedy decoding. Table~\ref{tab:prompt-settings} summarizes prompt formats and generation limits.

Ouro-1.4B and Ouro-2.6B use identical plain-text prompts and generation limits, without a chat template, for all four benchmarks. Ouro-2.6B evaluations use a maximum of 32 concurrent sequences with decode CUDA Graphs enabled and CPU offloading disabled. For Ouro-2.6B MATH500, the scorer truncates any appended new problem at a line-start Question: or Problem: heading before applying Math-Verify. The same extraction rule is applied to BF16 and every quantized configuration. The original generated outputs are retained for auditing. The prompt settings listed for Ouro-1.4B in Table~\ref{tab:prompt-settings} also apply to Ouro-2.6B. Huginn-3.5B uses \emph{You are a helpful assistant that can assist users with mathematical reasoning.} as the GSM8K system instruction and no separate system instruction for MATH500. The custom Ouro-1.4B MATH500 prompt requests step-by-step reasoning. MATH500 prompts request a final answer inside \texttt{\textbackslash boxed\{\}}, with Huginn-3.5B additionally instructed to use the \texttt{Final Answer:} format. Code prompts request Python code and provide an assistant response prefix with an open code fence. Generated outputs are cleaned before running the tests.

\subsection{Effective KV Bitwidth}
\label{subsec:effective-kv-bitwidth}
We measure KV cache storage using the effective bitwidth
$b_{\mathrm{eff}}$, defined as the total number of bits required to store
token-dependent quantized KV codes and their associated metadata, divided by
the number of original KV elements.
For a group of $L$ loops with equal KV sizes,
\begin{equation}
b_{\mathrm{eff}}
=
\frac{1}{L}
\sum_{r=1}^{L}
\left(
b_r + \frac{16}{g_r}
\right)
+
c_{\alpha},
\label{eq:effective-bitwidth}
\end{equation}
where $b_r$ and $g_r$ denote the quantization precision and group size,
respectively, for loop $r$.
The term $16/g_r$ accounts for the two FP8 metadata values---a scale and an
offset---stored for each quantization group.

For residual quantization with LS scaling, we additionally store one BF16 scaling coefficient for
each residual loop, separately for keys and values.
Its amortized storage cost is
\begin{equation}
c_{\alpha}
=
\frac{16(L-1)}{LD},
\end{equation}
where $D$ is the head dimension.
For direct quantization and OptR-H, which do not require these coefficients,
$c_{\alpha}=0$.

For Ouro-1.4B and Ouro-2.6B, with $L=4$ and $D=128$, LS coefficients contribute
$0.094$ additional bits per KV element.
The four configurations in Table~\ref{tab:main-benchmarks-reevaluation-method-rows-lowbit} are uniform INT2 with $g=32$, uniform INT2 with $g=16$, mixed INT2/4 with $g=32$, and mixed INT2/4 with $g=16$. For direct quantization and OptR-H, their effective bitwidths are approximately $2.5$, $3.0$, $3.0$, and $3.4$ bits, respectively. Residual quantization with LS scaling uses approximately $2.6$, $3.1$, $3.1$, and $3.5$ bits, respectively, due to the additional LS coefficients.
We include integer codes, scales, offsets, and LS coefficients in
$b_{\mathrm{eff}}$, but exclude fixed rotations and means, as well as
runtime padding and temporary buffers.

\section{Analysis Details}
\label{app:analysis-details}
\paragraph{Key prediction MSE (Figure~\ref{fig:residual-analysis}).}
The middle and right panels of Figure~\ref{fig:residual-analysis} use 512 GSM8K training examples, averaging per-example statistics with equal weights. The left panel follows the generation and timing protocol in Section~\ref{sec:throughput-experiments}. For the right panel, using one-based loop indices, let $K_{i,r,j}$ be the original key vector for problem $i$, loop $r\in\{1,2,3\}$, and layer--token--head position $j$. The Last-loop reference is the reconstructed anchor $R_{i,j}=\widehat K_{i,4,j}$. We measure
\begin{equation}
\mathrm{MSE}^{\mathrm{pred}}_{i,r}
=\frac{1}{N_{i,r}}\sum_j
\left\|K_{i,r,j}-\alpha_{i,r,j}R_{i,j}\right\|_2^2,
\end{equation}
where $N_{i,r}$ counts all key elements included in the sum. We compare $\alpha_{i,r,j}=1$ with LS scaling, $\alpha_{i,r,j}=\langle K_{i,r,j},R_{i,j}\rangle/\|R_{i,j}\|_2^2$, using the same reconstructed anchor under uniform INT2 or INT4 with group size $g=32$ and no rotation. This measures the residual before residual quantization, not the final key reconstruction error. We average MSE equally across the 512 problems.

\paragraph{Key reconstruction NMSE (Table~\ref{tab:reference-direction-inline}).}
\label{app:math500-reference-nmse}
For Table~\ref{tab:reference-direction-inline}, we collect BF16 post-RoPE keys using teacher forcing on the full prompts and gold solutions. First-loop and Last-loop use uniform INT2 with group size $g=32$, $\alpha_r=1$, and no rotation. Both policies use the same BF16 states and the packed cache writer and reader, without feedback from quantized attention. All 500 problems are included without token truncation. We compute $\|\widehat K-K\|_F^2/\max(\|K\|_F^2,10^{-30})$ per problem and loop over all tokens, layers, heads, and channels, then average equally across problems and multiply by $10^3$.

\section{Key Distributions Relative to the Last-loop Anchor}
\label{app:e1-transitions}

\begin{figure}[!htbp]
\centering
\includegraphics[width=0.85\linewidth]{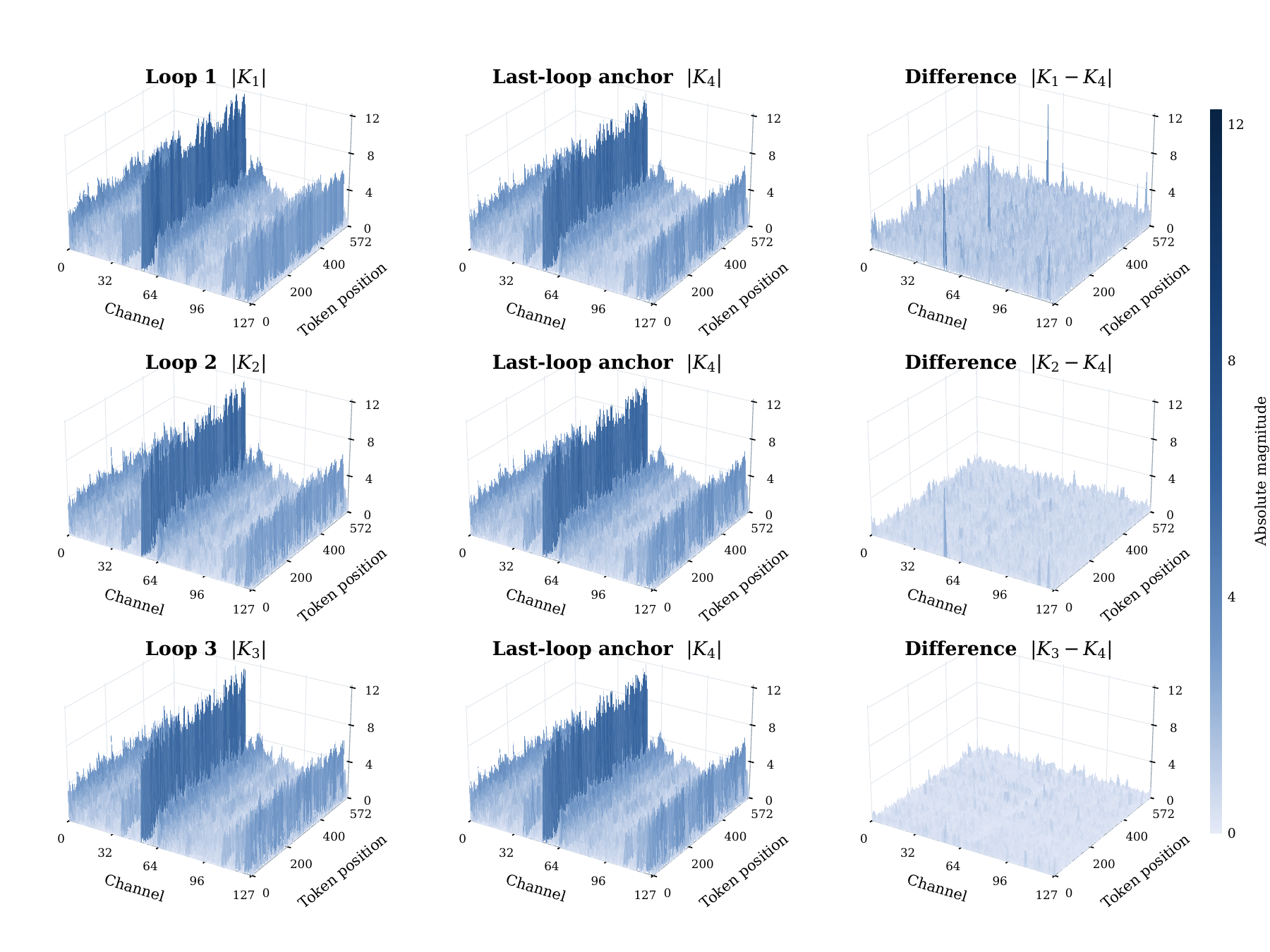}
\caption{Key distributions relative to the last-loop anchor. Rows correspond to loops 1, 2, and 3. Columns show $|K_r|$, $|K_4|$, and $|K_r-K_4|$ on shared scales.}
\label{fig:last-anchor-key-distributions}
\end{figure}

Using the same input as Figure~\ref{fig:e1_loop1_loop2}, Figure~\ref{fig:last-anchor-key-distributions} compares the keys from loops 1, 2, and 3 with the loop 4 anchor and their differences $K_r-K_4$. The data come from physical layer 3 and KV head 2 of Ouro-1.4B, with both indices starting at 1. Post-RoPE keys are collected during BF16 execution on a 573-token input comprising the fourth GSM8K training example and the preceding three examples as demonstrations. All tokens and 128 channels are shown, with token and channel indices starting at 0. Panels display absolute values on shared height and color scales. The differences are $K_r-K_4$ before LS adjustment or quantization.

\section{KV Reconstruction for Four Reference Choice Configurations}
\label{app:ouro-reconstruction}
Let $X_1,\ldots,X_4$ denote Ouro-1.4B's KV states at a fixed layer, token, and head. Residual quantization with LS scaling and shared OptR-H rotations reconstructs states in the rotated space $Z_r=UX_r$, yielding $\widehat X_r=U^\top\widehat Z_r$ in the original space. K and V use separate matrices $U$. Residual quantization without rotation corresponds to $U=I$.

Let $a$ denote the anchor loop and $p(r)$ the reference for loop $r$. The reconstructed anchor and residuals are
\begin{equation}
 A_a=Q(Z_a-\mu_a)+\mu_a,\qquad
 \widehat E_r=Q(Z_r-\alpha_r\widehat Z_{p(r)}-\mu_r)+\mu_r
\label{eq:ouro-general-codec}
\end{equation}
Here, $Q$ denotes quantization followed by reconstruction at the precision assigned to the corresponding loop, and $\alpha_r$ is the stored LS coefficient. The fixed key means $\mu_a$ and $\mu_r$ correspond to the rotated anchor and residuals, respectively. They are zero for V and for residual quantization without rotation. Each loop is reconstructed after recovering $A_a$ and $\widehat E_r$ from the cached codes, scales, and offsets.

\paragraph{Previous-loop.}
Loop 1 is the anchor, with $p(r)=r-1$. Reconstruction proceeds in the order $1\to2\to3\to4$.
\begin{equation}
 \widehat Z_1=A_1,\quad
 \widehat Z_2=\alpha_2 A_1+\widehat E_2,\quad
 \widehat Z_3=\alpha_3\widehat Z_2+\widehat E_3,\quad
 \widehat Z_4=\alpha_4\widehat Z_3+\widehat E_4.
\label{eq:ouro-previous-first}
\end{equation}
Reconstructing loop 4 requires the anchor followed by the residuals of loops 2, 3, and 4.

\paragraph{Next-loop.}
Loop 4 is the anchor, with $p(r)=r+1$. Reconstruction proceeds in the order $4\to3\to2\to1$.
\begin{equation}
 \widehat Z_4=A_4,\quad
 \widehat Z_3=\alpha_3 A_4+\widehat E_3,\quad
 \widehat Z_2=\alpha_2\widehat Z_3+\widehat E_2,\quad
 \widehat Z_1=\alpha_1\widehat Z_2+\widehat E_1.
\label{eq:ouro-previous-last}
\end{equation}
Reconstructing loop 1 requires the anchor followed by the residuals of loops 3, 2, and 1. This order describes reconstruction dependencies. The model still executes loops in the order $1\to2\to3\to4$.

\paragraph{First-loop.}
Loops 2, 3, and 4 directly reference loop 1, giving $p(r)=1$.
\begin{equation}
 \widehat Z_1=A_1,\quad
 \widehat Z_2=\alpha_2 A_1+\widehat E_2,\quad
 \widehat Z_3=\alpha_3 A_1+\widehat E_3,\quad
 \widehat Z_4=\alpha_4 A_1+\widehat E_4.
\label{eq:ouro-anchor-first}
\end{equation}
Each residual loop is reconstructed from $A_1$ and its own residual.

\paragraph{Last-loop.}
Loops 1, 2, and 3 directly reference loop 4, giving $p(r)=4$.
\begin{equation}
 \widehat Z_4=A_4,\quad
 \widehat Z_1=\alpha_1 A_4+\widehat E_1,\quad
 \widehat Z_2=\alpha_2 A_4+\widehat E_2,\quad
 \widehat Z_3=\alpha_3 A_4+\widehat E_3.
\label{eq:ouro-anchor-last}
\end{equation}

Each residual loop is reconstructed from $A_4$ and its own residual.

Thus, under First-loop and Last-loop, reconstructing any residual loop requires only two KV cache reads: the shared anchor and that loop's residual.
In contrast, Previous-loop and Next-loop require loading a chain of intermediate residuals, with the number of reads increasing with the distance from the anchor and reaching four reads for the farthest loop in a four-loop model.

\paragraph{Logical reference count.}
For a group of $L$ loops, consider the number of compressed components required to reconstruct every loop's KV. Neighboring-loop reconstruction uses $1,2,\ldots,L$ components depending on distance from the anchor, for a total of $L(L+1)/2$ accesses. Shared-anchor reconstruction uses one component for the anchor and two for each of the remaining $L-1$ loops, totaling $2L-1$. The ratio is therefore $L(L+1)/[2(2L-1)]$, giving $10/7\approx1.43$ for $L=4$ and $528/63\approx8.38$ for $L=32$. These are logical component accesses, not byte counts accounting for bitwidth, metadata size, cache hits, or actual HBM traffic. Huginn-3.5B's eight independent 4-loop groups require 80 and 56 accesses, respectively, retaining the ratio of 1.43.

\section{Huginn-3.5B Anchor Interval}
\label{app:huginn-anchor-interval}

\begin{table}[H]
\caption{Huginn-3.5B GSM8K accuracy (\%, 1,319 questions) and effective KV bitwidth $b_{\mathrm{eff}}$ across anchor intervals. Each group uses one final INT4 anchor. $2^{\times n}$ denotes $n$ consecutive INT2 loops. Group size $g$ applies to INT2, while INT4 uses $g=32$. Both residual quantization variants use LS scaling and the Last-loop policy. Effective bitwidth includes quantized values, scales, offsets, and LS coefficients, following Appendix~\ref{subsec:effective-kv-bitwidth}, and is the same with and without OptR-H.}
\label{tab:huginn-anchor-interval}
\centering
\small
\begin{tabular}{lcrrr}
\toprule
Schedule & $g$ & $b_{\mathrm{eff}}$ & w/o OptR-H & w/ OptR-H\\
\midrule
BF16 & -- & 16 & \multicolumn{2}{c}{42.84}\\
\midrule
\multirow{2}{*}{$8\times[2,2,2,4]$} & 16 & 3.5 & 41.70 & 42.38\\
 & 32 & 3.1 & 43.14 & 42.61\\
\midrule
\multirow{2}{*}{$4\times[2^{\times7},4]$} & 16 & 3.3 & 41.55 & 42.53\\
 & 32 & 2.9 & 41.93 & 42.15\\
\midrule
\multirow{2}{*}{$2\times[2^{\times15},4]$} & 16 & 3.3 & 41.93 & 42.68\\
 & 32 & 2.8 & 42.61 & 42.30\\
\bottomrule
\end{tabular}
\end{table}

The comparison keeps 32 total loops and varies schedules that use the final loop of each group as an INT4 anchor. Separate $96\times96$ rotations for K and V are learned for each group, layer, and head and shared within the group.

Although we use four-loop groups by default, grouping eight or sixteen loops retains comparable GSM8K accuracy with OptR-H while reducing the number of INT4 anchors. Increasing the group length from four to sixteen loops reduces effective bitwidth from 3.5 to 3.3 at $g=16$ and from 3.1 to 2.8 at $g=32$. The sixteen-loop configurations reduce logical KV storage by 79.7\% and 82.6\% relative to BF16, respectively. Accuracy changes from 42.38\% to 42.68\% and from 42.61\% to 42.30\%, respectively. Larger loop groups thus offer an additional option for reducing KV storage while retaining similar accuracy.

\clearpage

\section{Full KV Quantization Ablations}
\label{app:full-kv-ablations}

\begin{table}[!t]
\caption{\textbf{Scaling coefficient, reference, and rotation ablations.}
MATH500 accuracy (\%) on Ouro-1.4B with uniform INT2, uniform INT4, and mixed INT2/4. Uniform INT2 uses group size $g=16$, while uniform INT4 and mixed INT2/4 use $g=32$ in every loop.
For uniform precision, direct quantization is independent of the reference policy. For mixed INT2/4, Previous-loop and First-loop use $[4,2,2,2]$, while Next-loop and Last-loop use $[2,2,2,4]$. The direct and rotation-only baselines are listed separately for each mixed-precision schedule.
Residual variants compare different scaling coefficients and reference choices,
with OSCAR~\citep{zhou2026oscar} or OptR-H~\citep{yun2026optr} applied to residual quantization with LS scaling.
Blue row shading marks the selected Last-loop policy, blue column shading marks residual quantization with LS scaling and OptR-H, and the darker intersections indicate our final configuration.
BF16 accuracy is 75.0\%.
}
\label{tab:math500-uniform-analysis}

\centering
\footnotesize
\setlength{\tabcolsep}{3.5pt}
\renewcommand{\arraystretch}{1.05}

\begin{tabular}{lccccccc>{\columncolor{blue!6}}c}
\toprule
&
\multicolumn{4}{c}{No rotation}
& \multicolumn{2}{c}{+ OSCAR}
& \multicolumn{2}{c}{+ OptR-H}
\\
\cmidrule(lr){2-5}
\cmidrule(lr){6-7}
\cmidrule(lr){8-9}

Reference / Method
& Direct
& Residual
& Norm-ratio
& LS scaling
& Direct
& LS scaling
& Direct
& \cellcolor{blue!6}LS scaling
\\
\midrule

\multicolumn{9}{l}{\textbf{INT2}, \textit{$g=16$}} \\

Direct
& 54.20
& --
& --
& --
& 65.80
& --
& 64.20
& --
\\

Previous-loop
& --
& 69.20
& 73.00
& 71.20
& --
& 74.60
& --
& 72.20
\\

Next-loop
& --
& 71.60
& 67.60
& 73.20
& --
& 73.60
& --
& 72.80
\\

First-loop
& --
& 46.00
& 69.00
& 69.80
& --
& 71.20
& --
& 71.40
\\

\rowcolor{blue!6}
Last-loop
& --
& 69.20
& 68.80
& 67.80
& --
& 74.80
& --
& \cellcolor{blue!14}75.20
\\

\midrule

\multicolumn{9}{l}{\textbf{INT4}, \textit{$g=32$}} \\

Direct
& 74.20
& --
& --
& --
& 75.00
& --
& 75.20
& --
\\

Previous-loop
& --
& 74.40
& 76.00
& 76.20
& --
& 75.00
& --
& 74.40
\\

Next-loop
& --
& 76.00
& 76.20
& 74.60
& --
& 75.80
& --
& 74.80
\\

First-loop
& --
& 73.40
& 74.20
& 75.20
& --
& 75.00
& --
& 75.40
\\

\rowcolor{blue!6}
Last-loop
& --
& 75.80
& 74.00
& 74.20
& --
& 77.20
& --
& \cellcolor{blue!14}74.40
\\

\midrule

\multicolumn{9}{l}{\textbf{Mixed INT2/4}, \textit{$g=32$}} \\

Direct $[4,2,2,2]$
& 40.20
& --
& --
& --
& 61.20
& --
& 61.80
& --
\\

Direct $[2,2,2,4]$
& 47.40
& --
& --
& --
& 62.00
& --
& 63.00
& --
\\

Previous-loop
& --
& 66.20
& 74.20
& 73.20
& --
& 74.20
& --
& 72.80
\\

Next-loop
& --
& 71.40
& 75.40
& 74.00
& --
& 76.00
& --
& 75.20
\\

First-loop
& --
& 24.80
& 69.80
& 73.80
& --
& 74.60
& --
& 72.20
\\

\rowcolor{blue!6}
Last-loop
& --
& 70.20
& 74.60
& 73.80
& --
& 73.20
& --
& \cellcolor{blue!14}76.00
\\

\bottomrule
\end{tabular}
\end{table}

Table~\ref{tab:math500-uniform-analysis} provides a detailed comparison under uniform INT2, uniform INT4, and mixed INT2/4, covering the results in Tables~\ref{tab:math500-rotation-residual} and~\ref{tab:math500-reference-scaling}. It includes Norm-ratio scaling and both OSCAR and OptR-H rotations. For uniform precision, the direct and rotation-only baselines are independent of the reference choice and are listed once per precision. For mixed INT2/4, these baselines are listed separately for the $[4,2,2,2]$ and $[2,2,2,4]$ precision schedules. For a KV vector $X_r$ and its reconstructed reference $R_r$, we compare residual quantization with three scaling choices: $\alpha_r=1$, Norm-ratio scaling with $\alpha_r=\|X_r\|_2/\|R_r\|_2$, and LS scaling with $\alpha_r=\langle X_r,R_r\rangle/\|R_r\|_2^2$, which minimizes $\|X_r-\alpha_rR_r\|_2^2$. The LS + OSCAR and LS + OptR-H columns apply residual quantization with the same LS coefficient formula in the rotated space. Neither coefficient choice uniformly dominates the other in downstream accuracy.

\section{Rotation Calibration and Inference}
\label{app:rotation-details}
OSCAR and OptR-H both determine K and V rotations before inference, but use different criteria to obtain them.
This section describes how the rotation matrices are calibrated offline and how the resulting fixed matrices are applied during inference.

\subsection{Calibration}
Let $Q,K,V$ contain calibration tokens as rows for a fixed layer and head, and let $S=\operatorname{softmax}(QK^\top/\sqrt D)$ denote attention probabilities with a causal mask. The row-vector rotation corresponding to the main text's column-vector rotation $U$ is $T=U^\top$, applied to $X$ as $XT$. The rotations $T_K$ and $T_V$ are obtained separately. In \textsc{ResidualQuant}, the anchor and residuals in each four-loop group share the same matrices.

\paragraph{Quantized reference and residual targets.}
Q/K/V states are collected from the BF16 model, but residual references use quantized and reconstructed KV rather than original KV. With the last loop as anchor, $X_4$ is first quantized and reconstructed, after which each token's coefficient $\alpha_r=\langle X_r,\widehat X_4\rangle/\|\widehat X_4\|_2^2$ and residual $E_r=X_r-\alpha_r\widehat X_4$ are computed. The quantization targets are $X_4$ for the anchor and $E_r$ for residuals. Rotation-only baselines instead target original KV.

\paragraph{OSCAR.}
OSCAR \citep{zhou2026oscar} constructs rotations from covariance estimates of directions important for QK scores and weighted V outputs. Let $Z_V$ be the V quantization target, either original V or a residual computed with LS scaling, and $N_q$ the number of queries. Our adaptation uses the statistics and rotations
\begin{equation}
 C_K=\frac{Q^\top Q}{N_q},\qquad
 C_V=\frac{(SZ_V)^\top(SZ_V)}{N_q},\qquad
 C_X=B_X\Lambda_X B_X^\top,\quad T_X=B_XHP_X.
\end{equation}
Here, $B_X$ contains covariance eigenvectors, $H$ is a normalized Hadamard matrix, and $P_X$ is a bit-reversal permutation based on eigenvalue ranks. Eigenvectors align important directions, while Hadamard mixing and permutation redistribute them across groups. Shared rotations aggregate statistics over the corresponding loops using all calibration queries. Attention probabilities $S$ remain those computed from original Q/K, while LS variants use residuals as $Z_V$. Covariances and eigendecompositions are computed in FP64, without gradient-based rotation learning.

\paragraph{OptR-H.}
For OptR-H, mathematical reasoning evaluations share rotation matrices calibrated on the GSM8K training set, while code evaluations use matrices calibrated on the MBPP training set after excluding examples whose IDs overlap the evaluation set. Each domain uses 128 fit examples and 128 holdout examples with seed 0.

OptR \citep{yun2026optr} optimizes attention error induced by quantized KV. Our OptR-H implementation maintains orthogonality through $T=T_0\exp(A-A^\top)$ with Hadamard initialization $T_0=H$. Let $\widehat S=\operatorname{softmax}(Q\widehat K^\top/\sqrt D)$ and let $W_O$ denote the output projection for the head. The K and V objectives are
\begin{equation}
 \mathcal L_K=\overline{D}_{\mathrm{KL}}(S\|\widehat S)+\mathcal E((\widehat S-S)V),\qquad
 \mathcal L_V=\mathcal E(\widehat S(\widehat V-V)),\quad
 \mathcal E(\Delta)=\frac{\|\Delta W_O\|_F^2}{N_q d_{\mathrm{model}}}.
\end{equation}
Here, $\overline{D}_{\mathrm{KL}}$ is the mean query-wise KL divergence. The K rotation is selected first, and V rotation learning then fixes $\widehat S$ computed from reconstructed K. Output projection error is evaluated per head using $W_OW_O^\top$. Shared matrices minimize the mean loss across loops in the group.

For Ouro-1.4B, rotations and optimizer states use FP32 with Adam at learning rate 0.02 for 80 steps. Each step uses the last 64 queries from two calibration examples. Quantization uses straight-through gradients, while gradients are stopped through reconstructed references. Per-head matrices are selected by holdout loss.

\subsection{Inference}
After calibration, matrices are fixed and stored in FP16. Ouro-1.4B uses $128\times128$ rotations, while Huginn-3.5B uses $96\times96$ rotations matching its head dimension. Rotation multiplication uses FP16 inputs and FP32 accumulation. No further rotation training occurs during token generation.

\paragraph{Quantization and storage.}
Returning to column-vector notation, let $Z_r=UX_r$. Distinguish quantization into integer codes, $\mathcal Q_b$, from reconstruction using scales and offsets, $\mathcal D_b$. With fixed key mean $\mu_r$ in the rotated space, the last anchor and residuals are stored as
\begin{equation}
 c_4=\mathcal Q_4(Z_4-\mu_4),\quad
 A=\mathcal D_4(c_4)+\mu_4,\qquad
 c_r=\mathcal Q_2(Z_r-\alpha_r A-\mu_r)\quad(r=1,2,3).
\end{equation}
For V and OSCAR, $\mu_r=0$. The cache stores integer codes, group-wise FP8 scales and offsets, and BF16 LS coefficients for residuals. Matrices and fixed means are shared separately from the per-token cache. During decode, the current token participates in attention using its own BF16 KV. Compressed storage of its anchor and three residuals completes once the final loop's KV is available.

\paragraph{Reconstruction and attention.}
Rotated KV for past tokens is reconstructed as $\widehat Z_4=A$ and $\widehat Z_r=\alpha_r A+\mathcal D_2(c_r)+\mu_r$. Applying the same K rotation to the query gives $q^\top\widehat k_r=(U_Kq)^\top\widehat z_{K,r}$, avoiding inverse rotation of all keys. Values are inverse-rotated once after weighted accumulation, as $o=U_V^\top\sum_t p_t\widehat z_{V,r,t}$. The kernel reads compressed KV in tiles and computes anchor and residual contributions without creating a separate cache of fully reconstructed past KV. OSCAR and OptR-H thus supply fixed matrices to the same compressed attention path, allowing comparison of their calibration methods.

\section{Activation Residual Quantization}
\label{app:activation-details}
Table~\ref{tab:activation-extension} compares direct and residual activation quantization with LS scaling at W4A4. Weight and activation codes use INT4, while KV caches remain in BF16. The BF16 baseline in this section is evaluated alongside the activation experiments and is distinct from the KV experiment baseline.

\subsection{Baseline Methods and Implementation}
All W4A4 configurations use group size $g=32$ for both weights and activations, with BF16 KV caches.

\paragraph{Naive.}
Naive applies symmetric round-to-nearest quantization directly to weights and activations without a learned transformation or calibration. Its LS residual variant also requires no calibration.

\paragraph{FlatQuant.}
FlatQuant~\citep{sun2025flatquant} learns invertible affine transformations to flatten weight and activation distributions before quantization. The corresponding inverse transformation is folded into the weights, preserving the full-precision linear mapping before quantization. Kronecker-factorized matrices reduce transformation cost, while learnable per-channel scaling and weight and activation clipping further control outliers. These parameters are jointly calibrated by minimizing the output reconstruction error of each Transformer block. We use the authors' transformation, clipping, and calibration modules, adapted to the projection layouts of Ouro-1.4B and Huginn-3.5B. Each projection's transformation is shared across loops.

\paragraph{LoopQ.}
LoopQ~\citep{fang2026loopq} addresses loop-dependent distribution shifts and recursive error accumulation while retaining a shared quantized backbone. Loop-aware Activation Scaling (LAS) assigns activation ranges to individual loops. Selective Loop-aware Transformation (SLT) introduces loop-dependent transformations only for modules identified as sensitive to sharing by curvature-normalized inter-loop gradient variance. Lightweight transition adapters align each loop's quantized output before it enters the next loop. Trajectory-aware calibration jointly optimizes these components using full-precision logits, intermediate and final hidden states, and transition-alignment targets along the recurrent computation path.

In the absence of the authors' code, we reproduced LoopQ to the best of our ability by following the published methodology and matching the reported results across all benchmarks.

\subsection{Residual Formulation}
Let the row vector $x_r$ denote the current activation for a projection and $\widehat x_{r-1}$ the reconstructed reference from the previous loop. The first loop directly quantizes the full activation. Subsequent loops compute residuals using a token-wise scalar LS coefficient.
\begin{equation}
 \alpha_r=\frac{\langle x_r,\widehat x_{r-1}\rangle}{\|\widehat x_{r-1}\|_2^2},\qquad
 e_r=x_r-\alpha_r\widehat x_{r-1},\qquad
 \widehat x_r=\alpha_r\widehat x_{r-1}+Q(e_r).
\end{equation}
Here, $Q$ includes the method's transformation and INT4 quantization and reconstruction. We apply residual quantization with LS scaling to QKV, attention output, MLP gate/up, and down projections.

\subsection{Calibration}
For both models, FlatQuant direct calibration uses 128 Pile samples of 2,048 tokens each, seed 0, 15 epochs, and batch size 4. Its LS residual variant initializes from the corresponding direct artifact and undergoes a further 15 epochs of calibration through the residual computation path, using the same calibration tokens and batch size. Residuals are formed before the FlatQuant transformation, and the quantized residual is passed to the low-precision projection. Earlier quantized projections provide reconstructed references during residual calibration.

The Ouro-1.4B LoopQ LS residual variant reuses the previously calibrated direct bundle without additional training. For Huginn-3.5B, LoopQ direct calibration uses 1,024 Pile samples of up to 256 tokens each. The recorded calibration recipe specifies three layerwise epochs and a 500-step setting. The archived Ouro-1.4B direct bundle does not include its original calibration run record, so we do not infer that it used the Huginn-3.5B training schedule.

\subsection{Evaluation Protocol}
Table~\ref{tab:activation-extension} evaluates all 1,319 GSM8K, 500 MATH500, 164 HumanEval, and 378 MBPP problems. GSM8K uses flexible-extract exact match, with three-shot plain-text prompts for Ouro-1.4B and eight-shot prompts using the authors' chat template and system instruction for Huginn-3.5B. The generation limit is 1,024 tokens.

Ouro-1.4B MATH500 uses zero-shot prompts, a 2,048-token generation limit, and Math-Verify. Huginn-3.5B MATH500 uses the four-shot \texttt{minerva\_math500} task with plain-text prompts, no chat template or system instruction, a 2,048-token generation limit, and Minerva exact match. Huginn-3.5B evaluations split the 500 problems into disjoint shards of 167, 167, and 166 problems and aggregate correct-answer counts over all 500 problems. This activation experiment protocol differs from the Huginn-3.5B KV evaluation protocol in Appendix~\ref{app:experimental-setup}.

HumanEval and MBPP use zero-shot EvalPlus prompts and report base pass@1 on the original tests, with generation limits of 1,024 and 2,048 tokens, respectively. Model-specific prompts and scoring settings are shared between direct and LS residual variants. BF16 and Naive require no calibration, and all reported activation experiments retain BF16 KV caches.

\section{Additional Runtime Results}
\label{app:additional-runtime}

\paragraph{Prefill latency and GPU memory.}
Table~\ref{tab:prefill-offloading-additional} provides the complete 2K, 4K, 8K, and 16K results corresponding to the compact main-text Table~\ref{tab:prefill-offloading-main}. Experimental settings and the historical BF16 measurement protocol are identical to those described in Section~\ref{sec:prefill-offloading} and Table~\ref{tab:prefill-offloading-main}.
\begin{table}[!htbp]
\centering
\caption{\textbf{Prefill latency and GPU memory at all input lengths.}
Complete 2K, 4K, 8K, and 16K results under the same experimental settings as Table~\ref{tab:prefill-offloading-main}.
Latency is in ms and peak GPU memory in GiB. Parentheses report ratios relative to BF16 at each input length.}
\label{tab:prefill-offloading-additional}
\footnotesize
\setlength{\tabcolsep}{3pt}
\renewcommand{\arraystretch}{1.05}
\begin{adjustbox}{max width=\linewidth}
\begin{tabular}{@{}*{9}{c}@{}}
\toprule
& \multicolumn{2}{c}{2K} & \multicolumn{2}{c}{4K} & \multicolumn{2}{c}{8K} & \multicolumn{2}{c}{16K} \\
\cmidrule(lr){2-3}\cmidrule(lr){4-5}\cmidrule(lr){6-7}\cmidrule(lr){8-9}
Method & Latency & Peak & Latency & Peak & Latency & Peak & Latency & Peak \\
\midrule
BF16 & 133.4\,{\scriptsize ($1.00\times$)} & 4.37\,{\scriptsize ($1.00\times$)} & 262.8\,{\scriptsize ($1.00\times$)} & 5.98\,{\scriptsize ($1.00\times$)} & 587.0\,{\scriptsize ($1.00\times$)} & 9.20\,{\scriptsize ($1.00\times$)} & 1426.8\,{\scriptsize ($1.00\times$)} & 15.65\,{\scriptsize ($1.00\times$)} \\
None & 137.2\,{\scriptsize ($1.03\times$)} & 4.26\,{\scriptsize ($0.98\times$)} & 271.1\,{\scriptsize ($1.03\times$)} & 5.81\,{\scriptsize ($0.97\times$)} & 602.0\,{\scriptsize ($1.03\times$)} & 8.85\,{\scriptsize ($0.96\times$)} & 1427.4\,{\scriptsize ($1.00\times$)} & 14.97\,{\scriptsize ($0.96\times$)} \\
Serial & 186.1\,{\scriptsize ($1.40\times$)} & 3.18\,{\scriptsize ($0.73\times$)} & 361.3\,{\scriptsize ($1.37\times$)} & 3.65\,{\scriptsize ($0.61\times$)} & 772.5\,{\scriptsize ($1.32\times$)} & 4.54\,{\scriptsize ($0.49\times$)} & 1755.6\,{\scriptsize ($1.23\times$)} & 6.34\,{\scriptsize ($0.41\times$)} \\
\textbf{Overlap} & \textbf{137.4}\,{\scriptsize ($\boldsymbol{1.03\times}$)} & \textbf{3.55}\,{\scriptsize ($\boldsymbol{0.81\times}$)} & \textbf{272.0}\,{\scriptsize ($\boldsymbol{1.03\times}$)} & \textbf{4.36}\,{\scriptsize ($\boldsymbol{0.73\times}$)} & \textbf{602.5}\,{\scriptsize ($\boldsymbol{1.03\times}$)} & \textbf{5.19}\,{\scriptsize ($\boldsymbol{0.56\times}$)} & \textbf{1428.9}\,{\scriptsize ($\boldsymbol{1.00\times}$)} & \textbf{6.87}\,{\scriptsize ($\boldsymbol{0.44\times}$)} \\
\bottomrule
\end{tabular}
\end{adjustbox}
\end{table}

\begin{figure}[!htbp]
\centering
\includegraphics[width=0.8\linewidth]{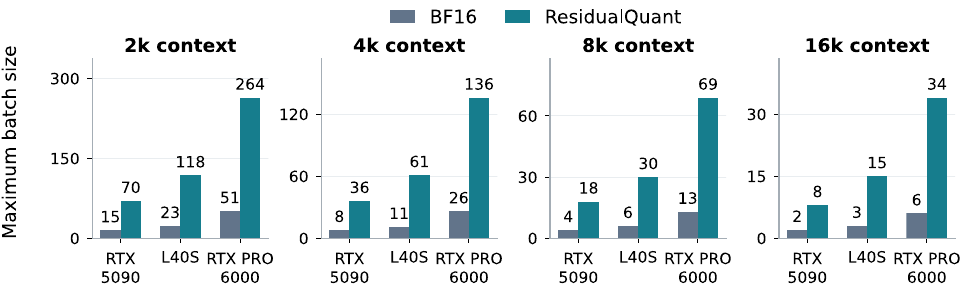}
\caption{\textbf{Max batch size at all input lengths.} BF16 and \textsc{ResidualQuant} on Ouro-1.4B on RTX 5090, L40S, and RTX PRO 6000 at 2k, 4k, 8k, and 16k contexts.}
\label{fig:max-batch-capacity-additional}
\end{figure}

\paragraph{Maximum batch size.}
Figure~\ref{fig:max-batch-capacity-additional} provides the complete 2k, 4k, 8k, and 16k results corresponding to Figure~\ref{fig:max-batch-capacity}, using the same GPUs and maximum feasible integer batch-size measurements.

\section{Detailed Related Work}
\label{app:detailed-related-work}

\paragraph{Looped Transformers.}
Reusing parameters across depth enables additional computation without a proportional increase in model size, as explored through cross-layer sharing and equilibrium formulations~\citep{lan2020albert,bai2019deq}. Adaptive computation further allocates this work according to the input, as in ACT, PonderNet, and Universal Transformers~\citep{graves2016act,banino2021pondernet,dehghani2019universal}. Recursive architectures extend these ideas through shared blocks, adaptive recursion, and memory of earlier loop states in RRT, MoR, and RecurTrace~\citep{bae2025relaxed,bae2025mor,wang2026recurtrace}. Such repeated computation supports length generalization and latent reasoning without intermediate token generation~\citep{fan2025length,saunshi2025latent}, with HRM, TRM, and Looped Flows exploring iterative refinement of internal states~\citep{wang2025hrm,jolicoeur2025trm,suleymanzade2026loopedflows}. To distinguish these benefits from simply spending more compute, SMELT studies sparse MoE looping under matched FLOPs, parameter, and KV cache budgets, demonstrating improved compute efficiency~\citep{wang2026smelt}.

\section{Comparison with FlashLoop}
\label{app:flashloop-comparison}

FlashLoop~\citep{yang2026flashloop} is concurrent work that combines INT4 KV quantization with sparse token updates and sparse attention for looped Transformer.
While its quantization method is motivated by a similar observation of inter-loop redundancy,
\textsc{ResidualQuant} improves both quantization
accuracy and runtime efficiency, enabling residual KV states to be reduced from INT4 to INT2.
Moreover, our quantization method is orthogonal to FlashLoop's sparse update
and attention techniques, allowing them to be combined for further
reductions in KV memory and computation while preserving higher accuracy.
We examine these aspects below.

\paragraph{Run-time efficiency.}
FlashLoop performs residual quantization using a Previous-loop reference,
where each loop is reconstructed from its predecessor.
This introduces a chain of reconstruction dependencies: recovering a later
loop requires loading the anchor together with multiple intermediate residuals
(Section~\ref{sec:reference-selection} and Appendix~\ref{app:ouro-reconstruction}).
In contrast, our Last-loop policy reconstructs every residual loop directly
from the final-loop anchor, reducing the required KV cache reads.
As shown in Figure~\ref{fig:residual-analysis}, under the same quantization setting, the
Previous-loop policy can only achieve 62\% of the decode throughput of the Last-loop policy due to larger memory traffic.

\paragraph{Low-precision accuracy.}
Reference choice, LS scaling, and rotation jointly affect accuracy at low precision.
As shown in Table~\ref{tab:math500-reference-scaling}, under uniform INT2 quantization with group size $g=16$ and without rotation, Previous-loop and Last-loop reconstruction with unit scaling both achieve 69.2\% MATH500 accuracy.
Combining Last-loop references with LS scaling and OptR-H achieves 75.2\% accuracy, compared with 64.2\% for OptR-H alone and 75.0\% for BF16 (Table~\ref{tab:math500-rotation-residual}).
These results support combining residual quantization with LS scaling and rotation to preserve accuracy close to BF16 at INT2.
In contrast, FlashLoop reports accuracy below 20\% when its residual quantization is reduced to INT2 in its ablation setting.

\begin{table}[H]
\caption{\textbf{Combining \textsc{ResidualQuant} with sparse KV updates and attention.}
MATH500 accuracy on Ouro-1.4B and average KV bitwidth during prefill and decode.
All methods use the same sparse KV update and attention mechanisms from FlashLoop.
For \textsc{ResidualQuant}, $g$ denotes the INT2 residual group size, while INT4 anchors use $g=32$.
Average bitwidth includes quantized values, scales, offsets, and LS coefficients, but excludes recent BF16 states.}
\label{tab:flashloop-math500-2}
\label{tab:selective-storage-math500-2}
\centering
\small
\setlength{\tabcolsep}{5pt}
\begin{tabular}{@{}lcrrr@{}}
\toprule
 & & \multicolumn{2}{c}{Average Bitwidth} & \\
\cmidrule(lr){3-4}
Method & BF16 KV tokens & Prefill & Decode & Accuracy (\%)\\
\midrule
BF16 + sparse & All & 9.40 & 16.00 & 76.0\\
\midrule
FlashLoop & Recent 64--127 & 2.64 & 4.50 & 72.8\\
\textsc{ResidualQuant} ($g=16$) + sparse & Recent 64 & 2.18 & 3.47 & \textbf{74.4}\\
\textsc{ResidualQuant} ($g=32$) + sparse & Recent 64 & \textbf{2.01} & \textbf{3.09} & 73.4\\
\bottomrule
\end{tabular}
\end{table}

\paragraph{Orthogonal integration with sparse updates and attention.}
\textsc{ResidualQuant} is orthogonal to FlashLoop's sparse KV update and
attention mechanisms.
Here, we further integrate \textsc{ResidualQuant} with FlashLoop's sparse update and
attention mechanisms to achieve higher accuracy with a smaller KV cache footprint.

We evaluate Ouro-1.4B on all 500 MATH500 problems using an
NVIDIA A100-SXM4-40GB, matching the GPU model reported by FlashLoop.
All methods use the same model revision, zero-shot plain prompts, greedy
decoding, seed 0, and a 2,048-token generation limit, with answers scored
using Math-Verify.
For FlashLoop, we directly use its public codebase\footnote{\url{https://github.com/Superone77/FlashLoop}} and default INT4
quantization with group size $g=64$.
For \textsc{ResidualQuant}, we use INT2/4 with INT2 group sizes
$g\in\{16,32\}$ and INT4 group size $g=32$.
FlashLoop's block-wise flushing retains between 64 and 127 recent tokens in
BF16; for \textsc{ResidualQuant}, we retain exactly 64 recent
tokens in BF16 for a fair comparison.
We directly apply the sparse update and attention mechanisms from the FlashLoop codebase to \textsc{ResidualQuant}.
Under sparsification, the final loop may not be retained for every token; in such cases, we use the last non-pruned loop as the INT4 anchor and represent the remaining retained loops with INT2 residuals.

As shown in Table~\ref{tab:flashloop-math500-2}, combining sparse updates and
attention with \textsc{ResidualQuant} improves both accuracy and KV storage
over FlashLoop.
With $g=16$, \textsc{ResidualQuant} reaches 74.4\% accuracy compared with
72.8\% for FlashLoop, while reducing average bitwidth from
2.64 to 2.18 for prefill and from 4.50 to 3.47 for decode.
With $g=32$, it further reduces bitwidth to 2.01 and 3.09,
respectively, while still improving accuracy to 73.4\%.
These results show that \textsc{ResidualQuant} provides a stronger quantization method even when combined with FlashLoop's sparse update and attention mechanisms, further improving the accuracy--memory tradeoff.

\end{document}